\documentclass[final,3p,times,twocolumn]{elsarticle}

\usepackage{amssymb}
\usepackage{algorithm}
\usepackage{algorithmic}
\usepackage{subfig}
\usepackage{graphicx}
\usepackage{amsmath}

\journal{Nuclear Physics B}

\begin{document}
\newcommand{\tabincell}[2]{\begin{tabular}{@{}#1@{}}#2\end{tabular}}

\begin{frontmatter}



\title{LTP: A New Active Learning Strategy for CRF-Based \\Named Entity Recognition}


\author{Mingyi Liu, Zhiying Tu, Tong Zhang, Tonghua Su, Zhongjie Wang}

\address{School of Computer Science and Technology, Harbin Institute of Technology, Harbin, China \\
    
    \{liumy, tzy\_hit, thsu, rainy\}@hit.edu.cn, muyejunzi@163.com}

\begin{abstract}
In recent years, deep learning has achieved great success in many natural language processing tasks including named entity recognition. The shortcoming is that a large amount of manually-annotated data is usually required. Previous studies have demonstrated that active learning could elaborately reduce the cost of data annotation, but there is still plenty of room for improvement. In real applications we found existing uncertainty-based active learning strategies have two shortcomings. Firstly, these strategies prefer to choose long sequence explicitly or implicitly, which increase the annotation burden of annotators. Secondly, some strategies need to invade the model and modify to generate some additional information for sample selection, which will increase the workload of the developer and increase the training/prediction time of the model. In this paper, we first examine traditional active learning strategies in a specific case of BiLstm-CRF that has widely used in named entity recognition on several typical datasets. Then we propose an uncertainty-based active learning strategy called Lowest Token Probability (LTP) which combines the input and output of CRF to select informative instance. LTP is simple and powerful strategy that does not favor long sequences and does not need to invade the model. We test LTP on multiple datasets, and the experiments show that LTP performs slightly better than traditional strategies with obviously less annotation tokens on both sentence-level accuracy and entity-level F1-score.
\end{abstract}

\begin{keyword}


active learning, named entity recognition, CRF
\end{keyword}

\end{frontmatter}


\section{Introduction}\label{sec:intro}
Over the past few years, papers applying deep neural networks (DNNs) to the task of named entity recognition (NER) have achieved noteworthy success \cite{chiu2016named, lample2016neural, limsopatham2016bidirectional}. However, under typical training procedures, the advantages of deep learning are established mostly relied on the huge amount of labeled data. When applying these methods on domain-related tasks, their main problem lies in their need for considerable human-annotated training corpus, which requires tedious and expensive work from domain experts. Thus, to make these methods more widely applicable and easier to adapt to various domains, the key is how to reduce the number of manually annotated training samples.

Active learning are designed to reduce the amount of data annotation. Unlike the supervised learning setting, in which samples are selected and annotated at random, the process of active learning employs one or more human annotators by asking them to label new samples that are supposed to be the most informative in the creation of a new classifier. The greatest challenge in active learning is to determine which sample is more informative. The most common approach is uncertainty sampling, in which the model preferentially selects samples whose current prediction is least confident.

Quite a lot of works have been done to reduce the amount of data annotation for NER tasks through active learning. However, these state-of-the-art approaches mainly face two problems. One of the problems is that they tend to choose the long sequences explicitly or implicitly, which will be an undesirable behavior when someone seeks to maximize performance for minimal cost annotation. Another problem is they may need to invade and modify the original model, which will increase the workload of the developer and increase the computing cost. \textbf{In this work}, we try to propose a simple but effective active learning strategy that does not prefer long sequence and does not need to invade original model.

When evaluating the effect of NER, most of the works only use the value of entity-level $F_1$ score. However, in some cases, this could be misleading, especially for languages that do not have a natural separator, such as Chinese. And the NER task is often used to support downstream tasks (such as QA, task-oriented dialogue), which prefer that all entities in the sentence are correctly identified. So \textbf{in this work}, we not only evaluate the entity-level $F_1$ score but also the sentence-level accuracy.  

We first experiment with the traditional uncertainty-based active learning algorithms, and then we proposed our own active learning strategy based on the lowest token probability with the best labeling sequence.  Experiments show that our selection strategy is superior to traditional uncertainty-based active selection strategies in multiple Chinese datasets and English datasets both in entity-level $F_1$ score and overall sentence-level accuracy. Finally, we make empirical analysis with different active selection strategies.

The remainder of this paper is organized as follows. In Section \ref{sec:related} we summarize the related works in named entity recognition and active learning. In section \ref{sec:model} we brief introduced the data representation and CRF. Section \ref{sec:strategies} describes in details the active learning strategies we propose. Section \ref{sec:experiments} describes the experimental setting, the datasets, and discusses the empirical results. And the last section is the conclusion.

\section{Related Work}\label{sec:related}
\subsection{Named entity recognition}
The framework of NER using deep neural network can be regarded as a composition of encoder and decoder. For encoders, there are many options. Collobert et al.\cite{collobert2011natural} first used convolutional neural network (CNN) as the encoder. Traditional CNN cannot solve the problem of long-distance dependency. In order to solve this problem, RNN\cite{nguyen2016toward}, BiLSTM\cite{huang2015bidirectional} , Dilated CNN\cite{strubell2017fast} and bidirectional Transformers\cite{devlin2018bert} are proposed to replace CNN as encoder. For decoders, some works used RNN for decoding tags \cite{nguyen2016toward, mesnil2013investigation}. However, most competitive approaches relied on CRF as decoder\cite{lample2016neural, yang2016multi}.

\subsection{Active learning}
Active learning strategies have been well studied \cite{dasgupta2005analysis,awasthi2014power}, \cite{shen2017deep}. These strategies can be grouped into following categories: \textit{Uncertainty sample} \cite{lewis1994heterogeneous,culotta2005reducing,scheffer2001active,kim2006mmr}, \textit{query-by-committee}\cite{seung1992query,vandoni2019evidential}, \textit{information density}\cite{wei2015submodularity}, \textit{fisher information}\cite{settles2008analysis}. There were some works that compared the performance of different types of selection strategies in NER/sequence labeling tasks with CRF model \cite{CHEN201511, Marcheggiani2014AnEC,settles2008analysis,10.1007/978-3-319-77113-7_3}. These results show that, in most case, uncertainty-based methods perform better and cost less time. 

However, we found that these studies are mainly based on English datasets, and don't pay much attention to Chinese datasets. Additionally, traditional uncertain-based strategies always choose long sequence explicitly or implicitly, which significantly increases the burden on the annotators. And some strategies \cite{siddhant2018deep} invade the model and let the model perform additional tasks for sample selection. So, in this work we proposed a new active learning strategy that does not favor long sequences and does not need to invade the model.

\section{NER Model}\label{sec:model}
\begin{table*}[!tbhp]
 \centering
  \caption{Example of data representation. [PAD] tag are not shown.}
  \label{tab:data_rep}
  \begin{tabular}{l|lllllllll}
    \hline
Sentence &       & Trump & was & born & in & the   & United & States &       \\ \hline
Tag      & [CLS] & B-PER & O   & O    & O  & B-LOC & I-LOC  & I-LOC  & [SEP] \\ \hline
\end{tabular}
\end{table*}
 \subsection{Data Representation}
 We represent each input sentence following Bert format; Each token in the sentence is marked with BIO scheme tags. Special $[CLS]$ and $[SEP]$ tokens are added at the beginning and the end of the tag sequence, respectively. $[PAD]$ tokens are added at the end of sequences to make their lengths uniform. The formatted sentence in length $N$ is denoted as $\mathbf{x} = <x_1, x_2, \dots , x_N>$, and the corresponding tag sequence is denoted as $\mathbf{y} = <y_1, y_2, \dots , y_N>$.

\subsection{CRF Layer} 
CRF are statistical graphical models which have demonstrated state-of-art accuracy on virtually all of the sequence labeling tasks including NER task.  Particularly, we use linear-chain CRF that is a popular choice for tag decoder, adopted by most DNNs for NER.

A linear-chain CRF model defines the posterior probability of $\mathbf{y}$ given $\mathbf{x}$ to be:
\begin{equation} \label{crf}
    P(\mathbf{y}|\mathbf{x};A) = \frac{1}{Z(\mathbf{x})}\exp \left(P(y_1;\mathbf{x}_1) + \sum_{k=1}^{n-1} P(y_{k+1};\mathbf{x}_{k+1}) + A_{y_k,y_{k+1}} \right)
\end{equation}
where $Z(\mathbf{x})$ is a normalization factor over all possible tags of $\mathbf{x}$, and $P(y_k;\mathbf{x}_k)$ indicates the probability of taking the $y_k$ tag at position $k$ which is the output of the previous DNN layer, such as bilstm, softmax. $A$ is a parameter called a transfer matrix, which can be set manually or by model learning. In our experiment, we let the model learn the parameter by itself. $A_{y_k,y_{k+1}}$ means the probability of a transition from tag states $y_k$ to $y_{k+1}$. We use $\mathbf{y}^{*}$ to represent the most likely tag sequence of $\mathbf{x}$:
\begin{equation}
    \mathbf{y}^{*} = \arg\max_{\mathbf{y}} P(\mathbf{y}|\mathbf{x})
\end{equation}
The parameters $A$ are learnt through the maximum log-likelihood estimation, that is to maximize the log-likelihood function $\ell$ of training set sequences in the labeled data set $\mathcal{L}$:
\begin{equation}
    \ell(\mathcal{L};A) =  \sum_{l=1}^{L}\log P(\mathbf{y}^{(l)}|\mathbf{x}^{(l)};A) 
\end{equation}
where $L$ is the size of the tagged set $\mathcal{L}$.

\section{Active Learning Strategies} \label{sec:strategies}
The biggest challenge in active learning is how to select instances that need to be manually annotated. A good selection strategy $\phi (\mathbf{x})$, which is a function used to evaluate each instance $\mathbf{x}$ in the unlabeled pool $\mathcal{U}$, will select the most informative instance $\mathbf{x}$.

\begin{algorithm}[htbp]
\caption{Pool-based active learning framework}
\label{alg:pool_based}
\begin{algorithmic} 
\REQUIRE Labeled data set $\mathcal{L}$, \\
        \quad \quad \quad unlabeled data pool $\mathcal{U}$, \\
        \quad \quad \quad selection strategy $\phi(\cdot)$, \\
        \quad \quad \quad query batch size $B$
\WHILE{\textbf{not} reach stop condition}
\STATE $//$ Train the model using labeled set $\mathcal{L}$
\STATE $train(\mathcal{L})$;
\FOR{$b=1$ to $B$}
\STATE $//$select the most informative instance
\STATE $\mathbf{x}^{*} = \arg\max_{\mathbf{x}\in\mathcal{U}}\phi(\mathbf{x})$
\STATE $\mathcal{L} = \mathcal{L} \cup <\mathbf{x}^{*}, label(\mathbf{x}^{*})>$
\STATE $\mathcal{U} = \mathcal{U} - \mathbf{x}^{*}$
\ENDFOR
\ENDWHILE
\end{algorithmic}
\end{algorithm}

Algorithm \ref{alg:pool_based} illustrate the entire pool-based active learning process. In the remainder of this section, we describe various query strategy formulations of $\phi(\cdot)$ in detail.

\subsection{Token-based (Local) Strategies}\label{sec:mtp}
The token-based strategy treats the labeling sequence as a set of isolated tokens, and evaluates uncertainty by aggregating the information of these tokens.

\textbf{Minimum Token Probability (MTP)} selects the most informative tokens, regardless of the assignment performed by CRF. This strategy greedily samples tokens whose highest probability among the labels is lowest:
\begin{equation}
	\phi^{MTP}(\mathbf{x}) =  1 - \min_{i}\max_{j}P(y_i=j|\mathbf{x}_i; A)
\end{equation}

where $P(y_i=j)$ is the probability that $j$ is the label at position $i$ in the sequence.

\textit{Entropy} is a popular measure of informativeness. The entropy of a discrete random variable $Y$ can be represented by $H(Y)=-\sum_{i}P(y_i)\log P(y_i)$, and means the information needed to "encode" the distribution of outcomes for $Y$. \textbf{Token Entropy (TE)} is a way to use the entropy of model's posteriors over its labeling:
\begin{equation}
	\phi^{TE} = -\frac{1}{N}\sum_{i=1}^N\sum_{j=1}^M P(y_i=j|\mathbf{x}_i; A)\log P(y_i=j|\mathbf{x}_i; A)
\end{equation}
where $N$ is the length of $\mathbf{x}$ without [PAD], $j$ ranges over all possible token labels.

Settles \cite{settles2008analysis} argue that querying long sequences should not be explicitly discouraged, if in fact they contain more information. They extend \textbf{TE} into \textbf{Maximum Token Entropy (MTE)}:
\begin{equation}
	\phi^{MTE}(\mathbf{x}) = N \times \phi^{TE}(\mathbf{x})
\end{equation}

\subsection{Sentence-based (Global) Strategies}
Different from token-based strategies, sentence-based strategies treat labeling sequence $\mathbf{y}$ as whole. Most of these strategies have high complexity or require intrusive models.

Culotta and McCallum \cite{culotta2005reducing} employ a simple uncertainty-based strategy for sequence models called least confidence (LC), which sort examples in ascending order according to the probability assigned by the model to the most likely sequence of tags:
\begin{equation}
    \phi^{LC}(\mathbf{x}) = 1 - P(\mathbf{y}^{*}|\mathbf{x};A)
\end{equation}
This confidence can be calculated using the posterior probability given by Equation \ref{crf}. Preliminary analysis revealed that the LC strategy prefer selects longer sentences:
\begin{small}
\begin{equation}\label{equ:lc}
    P(\mathbf{y}^{*}|\mathbf{x};A) \varpropto \exp \left(P(y^*_1;\mathbf{x}_1) + \sum_{k=1}^{n-1} P(y^*_{k+1};\mathbf{x}_{k+1}) + A_{y^*_k,y^*_{k+1}} \right)
\end{equation}
\end{small}
Since Equation \ref{equ:lc} contains summation over tokens, LC method naturally favors longer sentences. Although the LC method is very simple and has some shortcomings, many works have proved the effectiveness of the method in sequence labeling tasks.

Scheffer et al. \cite{scheffer2001active} propose a method called \textbf{Margin}, which queries samples with the smallest margin between the posteriors for its two most likely annotations:
\begin{equation}
	\phi^{M}(\mathbf{x}) = -(P(\mathbf{y}^*_1)|\mathbf{x}; A) - P(\mathbf{y}^*_2|\mathbf{x};A))
\end{equation}
where, $\mathbf{y}^*_1$ and $\mathbf{y}^*_2$ are the first and second most likely tag sequence of $\mathbf{x}$. \textbf{Margin} requires the model to calculate the unnesseary second most likely tag sequence.

Different from \textbf{TE} and \textbf{TTE}, \textbf{Sequence Entropy (SE)} considers the entropy of the sequence instead of the entropy of the token:
\begin{equation}
	\phi^{SE}(\mathbf{x}) = -\sum_{\hat{\mathbf{y}}}P(\hat{\mathbf{y}}|\mathbf{x}; A)\log P(\hat{\mathbf{y}}|\mathbf{x}; A)
\end{equation}
where $\hat{\mathbf{y}}$ ranges all over possible tag sequences for $\mathbf{x}$. This calculation cost will increase exponentially with the length of $\mathbf{x}$ and the number of tag categories.

The most recent uncertainty-based selection strategy is called \textbf{Bayesian Active Learning by Disagreement (BALD)}\cite{siddhant2018deep,gal2017deep}. BALD	 measures the uncertainty of the sample by observing the changes in the forward propagation result of the sample through multiple random dropouts\cite{gal2016theoretically}. Let $\tilde{\mathbf{y}}^1, \tilde{\mathbf{y}}^2, \dots, \tilde{\mathbf{y}}^T$ represent the result from apply $T$ independently sampled dropout masks:
\begin{equation}
	\phi^{BALD}(\mathbf{x}) = 1 - \frac{\max_{\tilde{\mathbf{y}}}count(\tilde{\mathbf{y}})}{T}
\end{equation}
where $count(\tilde{\mathbf{y}})$ means the number of occurrences of $\tilde{\mathbf{y}}$ in $\tilde{\mathbf{y}}^1, \tilde{\mathbf{y}}^2, \dots, \tilde{\mathbf{y}}^T$. Normally, the value of $T$ is $100$. BALD will cost a lot of time on repeating forward propagation when the data pool is large.

\subsection{Lowest Token Probability (LTP)}
Unlike existing strategies, we believe that local information and global information have their own advantages, and the two can complement each other. We look for the most probable sequence assignment (global), and hope that each token (local) in the sequence has a high probability.
\begin{equation}\label{eq:ltp}
    \phi^{LTP}(\mathbf{x}) = 1 - \min_{y_i^* \in \mathbf{y}^*}P(y^*_i|\mathbf{x}_i;A)
\end{equation}
We proposed our select strategy called \textbf{Lowest Token Probability (LTP)}, which selects the tokens whose probability under the most likely tag sequence $\mathbf{y}^*$ is lowest. It is not difficult to infer from the formulation that \textbf{LTP} utilizes global and local information, and implicitly implements \textbf{Margin} but does not require additional calculations\footnote{If there is a small probability token in the best sequence, then there is a high probability that the margin between 1st best sequence and 2nd best sequence is small}.

\begin{table*}[htb]
\centering
  \caption{Qualitative comparison of uncertainty-based active learning strategies}
  \label{tab:strategy_comp}
\begin{tabular}{l|llllllll}\hline

                                                                         & MTP & LC & TE & TTE & LTP & Margin & SE & BALD \\ \hline
Local(Token) Information  & $\surd$   &    & $\surd$  & $\surd$   & $\surd$   &        &    &      \\
Global(Sentence) Information  &     & $\surd$  &    &     & $\surd$   & $\surd$      & $\surd$  & $\surd$    \\
\begin{tabular}[c]{@{}l@{}}Favor long sequence\\ explicitly\end{tabular} &     & $\surd$  &    & $\surd$   &     &        &    &      \\
Invade model                                                             &     &    &    &     &     & $\surd$      & $\surd$  &      \\
Additional compute                                                       &     &    &    &     &     & $\surd$      & $\surd$  & $\surd$    \\ \hline
\end{tabular}
\end{table*}

Table \ref{tab:strategy_comp} compares all the uncertainty-based active learning strategies mentioned in this section. Strategies that do not need to invade the model and do not require additional calculations are selected as the comparison method of our strategies.

\section{Experiments}\label{sec:experiments}

\subsection{Datasets}
\begin{figure}[htbp]

\centering
\subfloat[]{\includegraphics[width=0.3\linewidth]{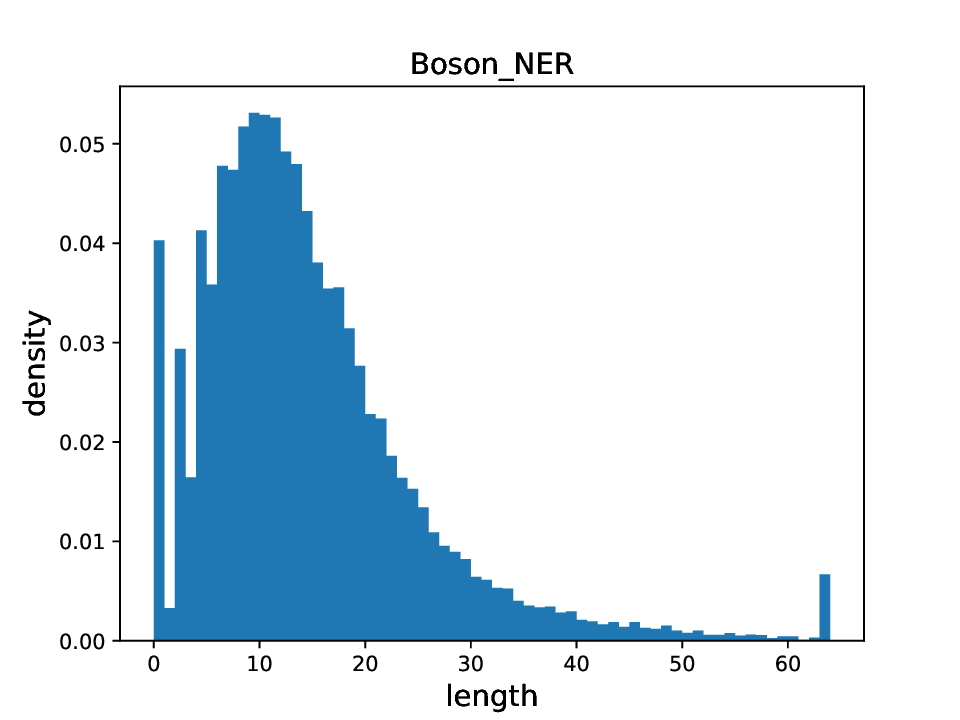}}\hfil
\subfloat[]{\includegraphics[width=0.3\linewidth]{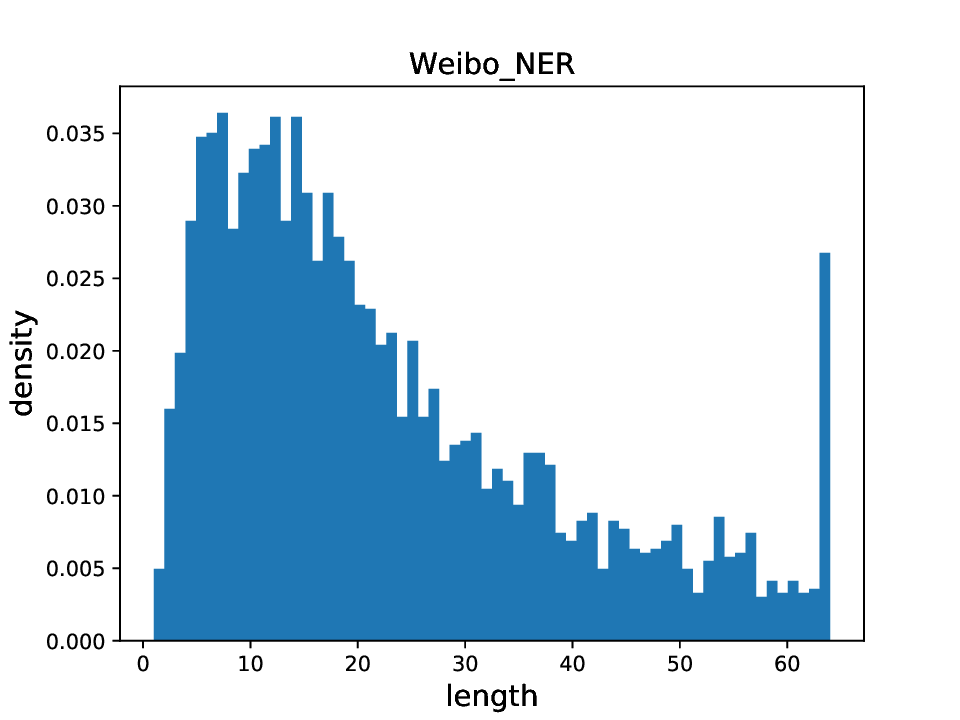}}\hfil 
\subfloat[]{\includegraphics[width=0.3\linewidth]{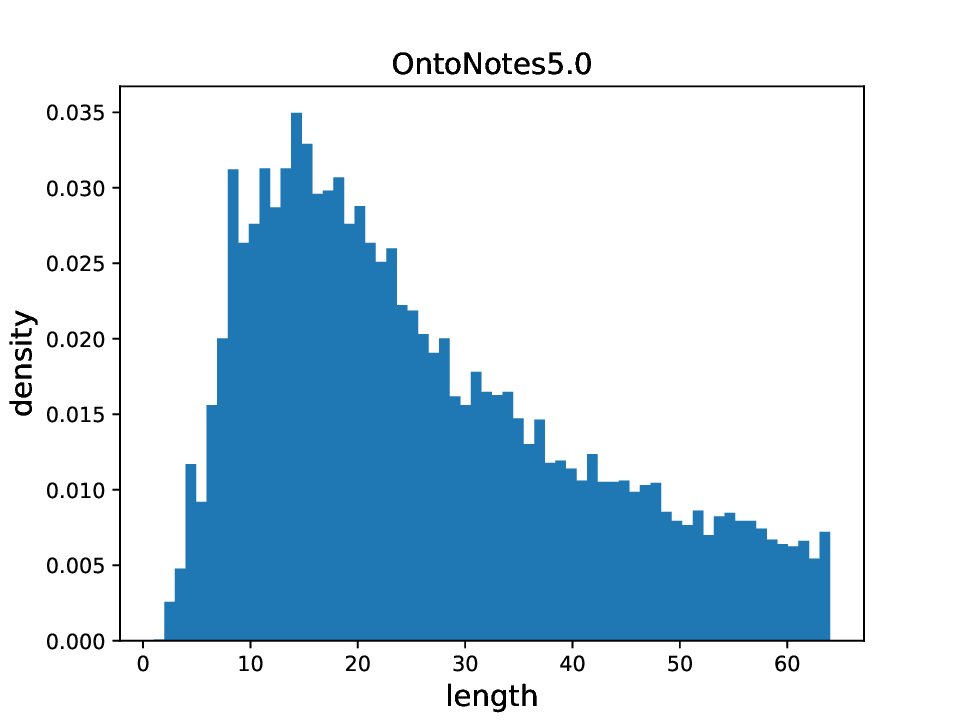}} 

\subfloat[]{\includegraphics[width=0.3\linewidth]{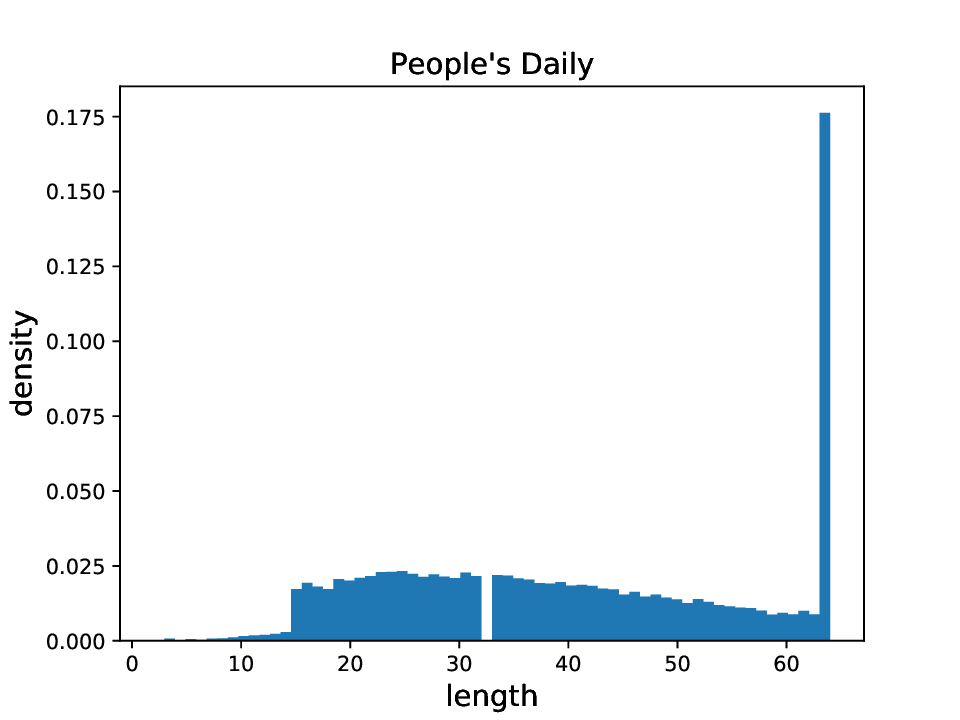}}\hfil   
\subfloat[]{\includegraphics[width=0.3\linewidth]{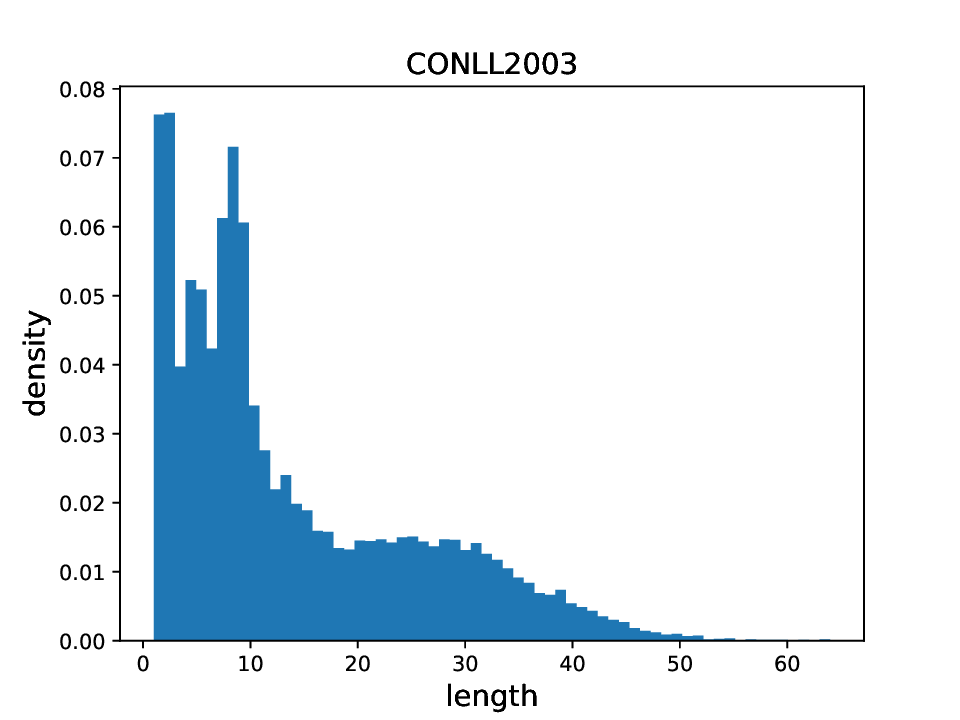}}\hfil
\subfloat[]{\includegraphics[width=0.3\linewidth]{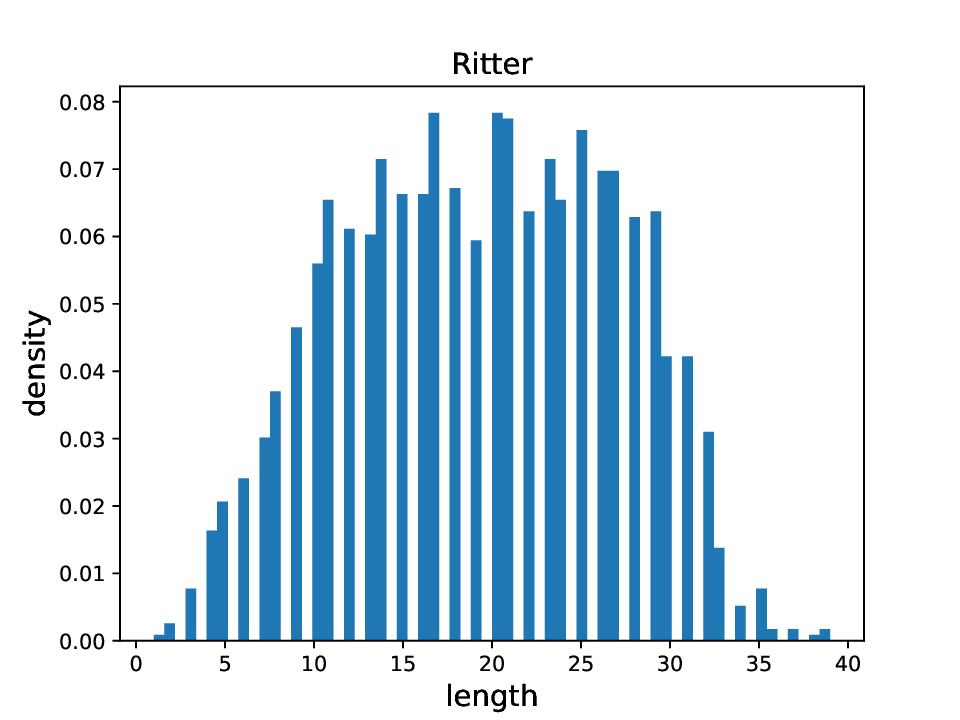}}
\caption{Distribution of sample lengths on different datasets}\label{fig:len_density}
\end{figure}
We have experimented and evaluate the active learning strategies mentioned on Section \ref{sec:strategies} on four Chinese datasets and two english datasets. \textit{People's Daily} is a collection of newswire article annotated with 3 balanced entity types; \textit{Boson\_NER\footnote{https://bosonnlp.com/resources/BosonNLP\_NER\_6C.zip}} is a set of online news annotations published by bosonNLP, which contains 6 entity types; \textit{Weibo\_NER}\cite{peng2015ner, peng2016improving} is a collection of short blogs posted on Chinese social media Weibo with 8 extremely unbalanced entity types; \textit{OntoNotes-5.0}\cite{weischedel2012ontonotes} Chinese dataset used in this paper is a collection of broadcast news articles, which contains 18 entity types. \textit{CONLL2003}\cite{sang2003introduction} is a well known english dataset consists of Reuters news stories between August 1996 and August 1997, which contains 4 different entity types; \textit{Ritter}\cite{Ritter11} is a english dataset consist of tweets annotated with 10 different entity types. All datasets are formatted in the "BIO" sequence representation. In order to be able to perform batch training, the length of all samples is limited to $64$. Those samples that were originally longer than $64$ will be split according to commas or directly truncated to meet the length requirement.

\begin{figure}[htbp]
\centering
\subfloat[]{\includegraphics[width=0.3\linewidth]{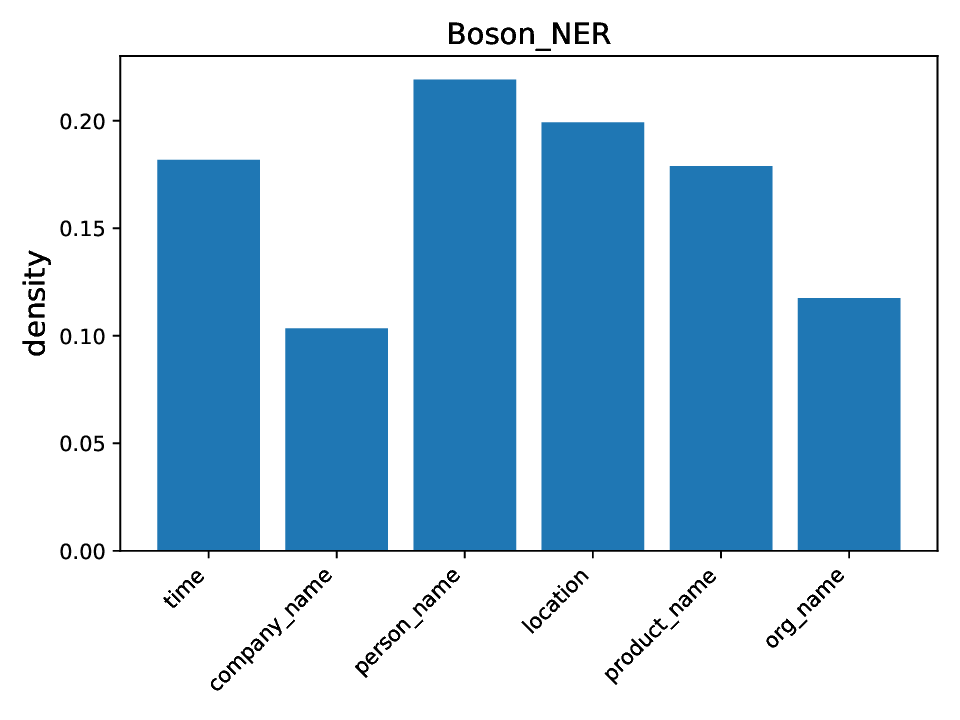}}\hfil
\subfloat[]{\includegraphics[width=0.3\linewidth]{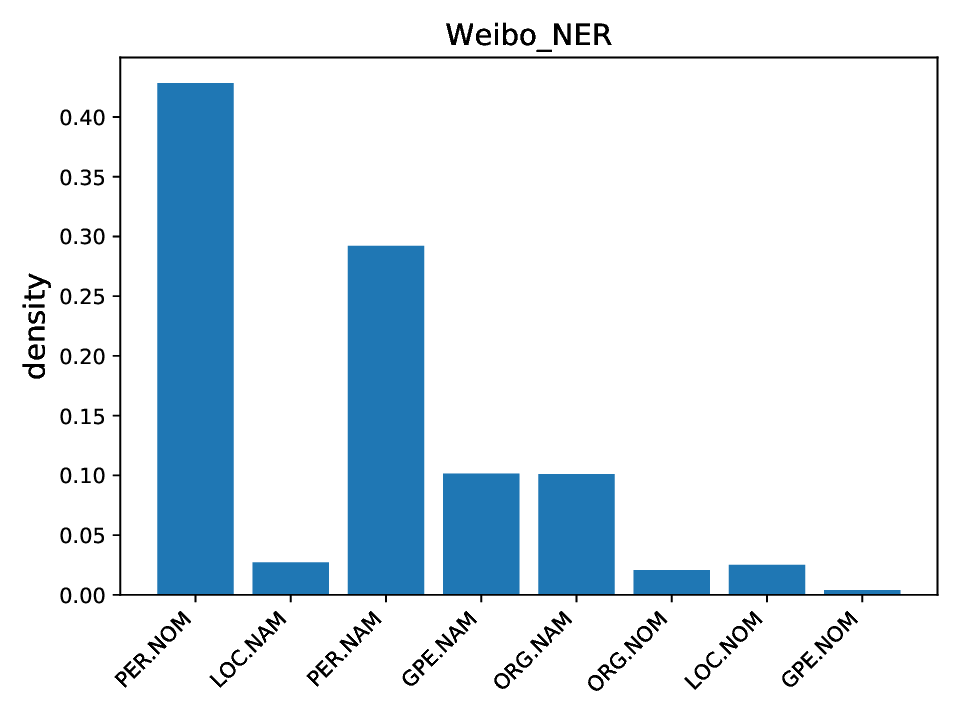}}\hfil 
\subfloat[]{\includegraphics[width=0.3\linewidth]{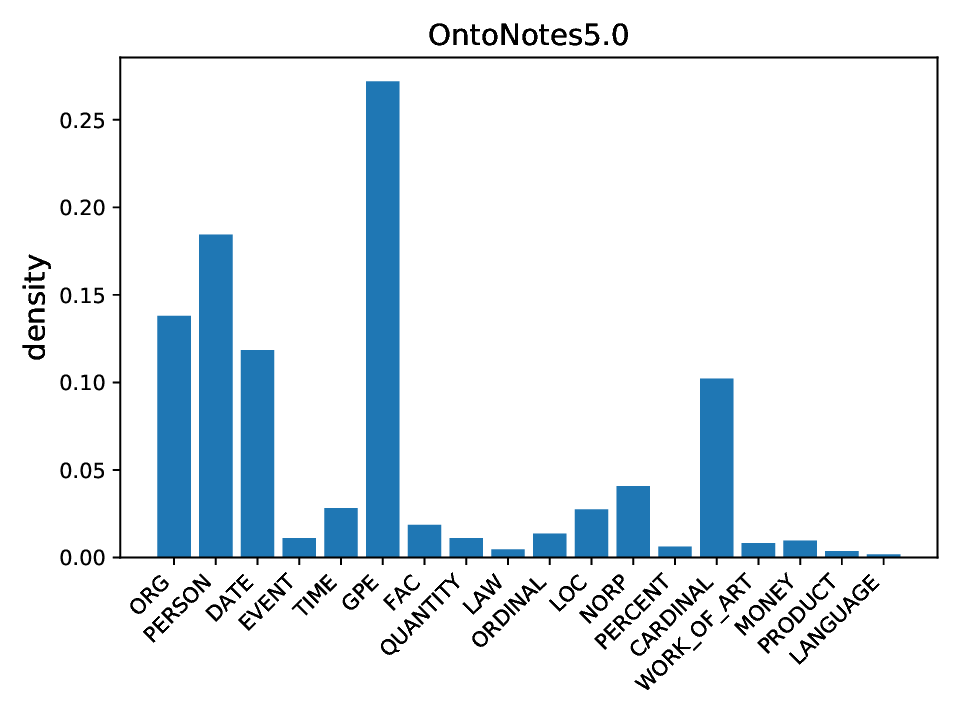}} 

\subfloat[]{\includegraphics[width=0.3\linewidth]{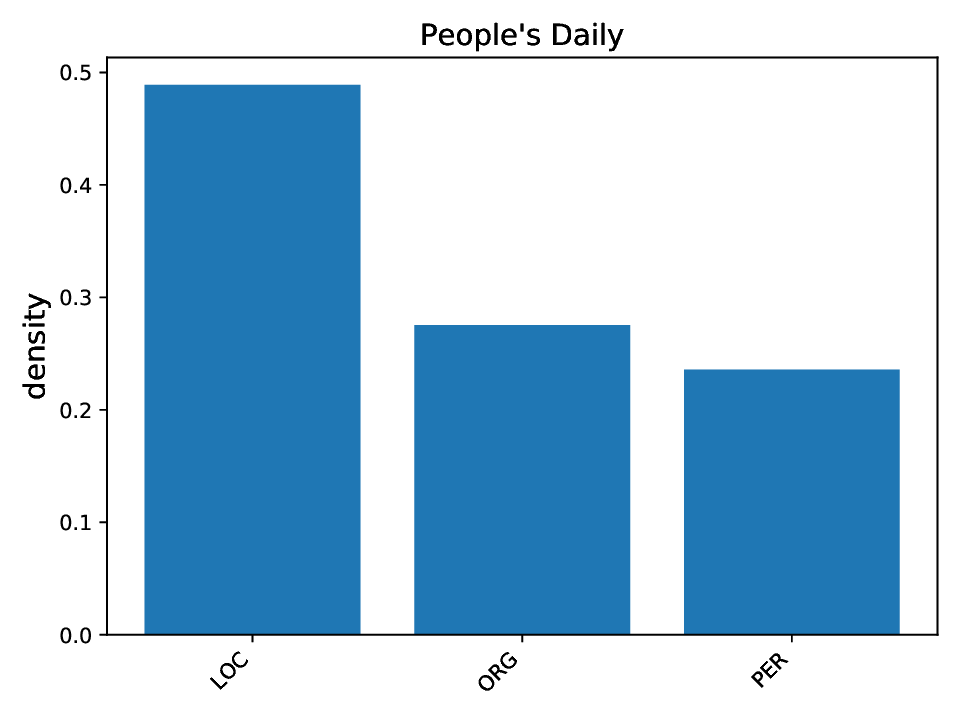}}\hfil   
\subfloat[]{\includegraphics[width=0.3\linewidth]{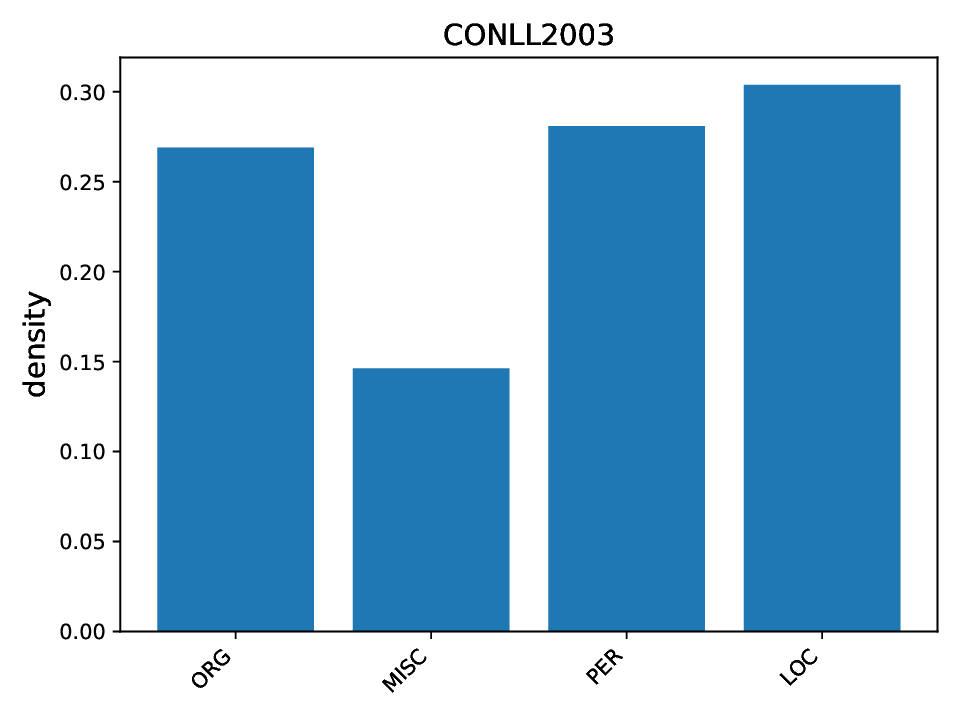}}\hfil
\subfloat[]{\includegraphics[width=0.3\linewidth]{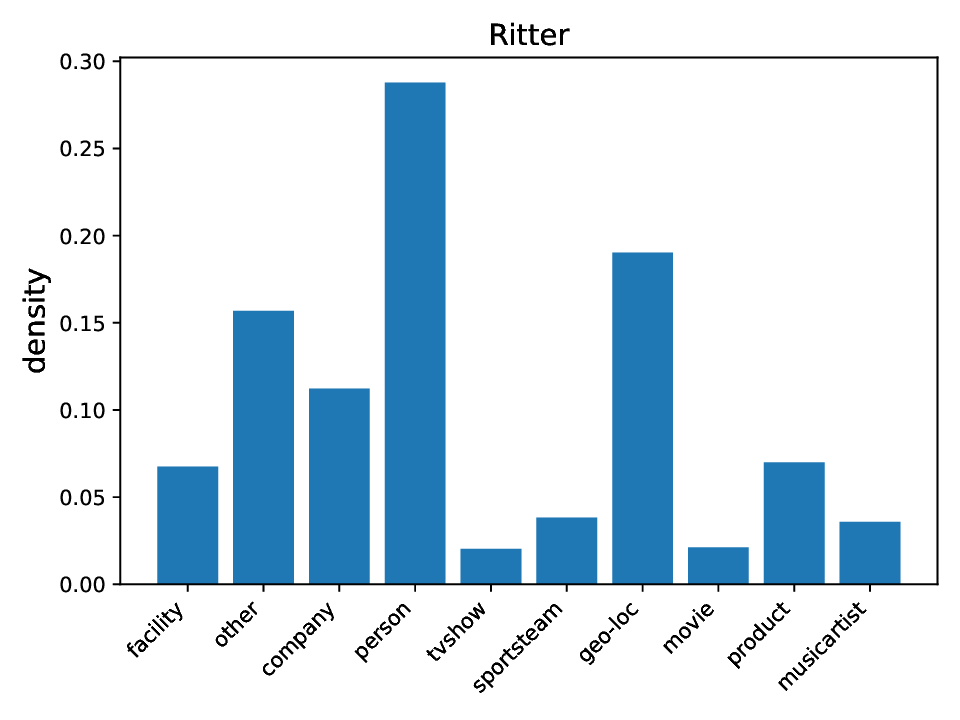}}
\caption{Distribution of entity types on different datasets}\label{fig:entity_density}
\end{figure}
Table \ref{tab:corpus} shows some statistics of the datasets in terms of dimensions, number of entity types, distribution of the labels, etc. Figure \ref{fig:len_density} gives the distribution of sample lengths on different datasets. Figure \ref{fig:entity_density} presents the distribution of entity types on different datasets. According to the description and statistical information of these datasets, we can conclude that these datasets are 6 datasets with obvious differences in language, text style, entity distribution, length distribution, data magnitude, etc.

\begin{table*}[htbp]
    \centering
     \caption{Training(Testing) Data Statistics. \#S is the number of total sentences in the dataset, \#T is the number of tokens in the dataset, \#E is the number of entity types, ASL is the average length of a sentence, ASE is the average number of entities in a sentence, AEL is the average length of a entity, \%PT is the percentage of tokens with positive label,\%AC is the percentage of a sentences with more than one entity, \%DAC is the percentage of sentences that have two or more entities. English datasets are marked in bold.}
    \label{tab:corpus}
\begin{tabular}{lrrrrrrrrr} 
\hline
corpus &    \#S &   \#T &   \#E &    ASL &  ASE &  AEL &   \%PT &  \%AC &  \%DAC  \\\hline 
Boson\_NER &   \tabincell{c}{27350\\ (6825)} & \tabincell{c}{409830\\ (99616)} &   \tabincell{c}{6\\ (6)} &  \tabincell{c}{14.98\\ (14.59)} &    \tabincell{c}{0.67\\ (0.67)} &  \tabincell{c}{3.93\\ (3.87)} &  \tabincell{c}{17.7\%\\ (17.8\%)} & \tabincell{c}{41.8\%\\ (41.8\%)} &  \tabincell{c}{14.7\%\\ (14.8\%)} \\
Weibo\_NER &   \tabincell{c}{3664\\ (591)} &   \tabincell{c}{85571\\ (13810)} &    \tabincell{c}{8\\ (8)} &  \tabincell{c}{23.35\\ (23.36)} &   \tabincell{c}{0.62\\ (0.66)} &  \tabincell{c}{2.60\\ (2.60)} &  \tabincell{c}{6.9\%\\ (7.3\%)} &   \tabincell{c}{33.6\%\\ (36.3\%)} &  \tabincell{c}{14.8\%\\ (17.7\%)} \\
\tabincell{c}{OntoNotes5.0\\ (bn-zh)} &   \tabincell{c}{13798\\ (1710)} & \tabincell{c}{362508\\ (44790)} & \tabincell{c}{18\\ (18)} &    \tabincell{c}{26.27\\ (26.19)} & \tabincell{c}{1.91\\ (1.99)} &    \tabincell{c}{3.14\\ (3.07)} & \tabincell{c}{22.8\%\\ (23.4\%)} & \tabincell{c}{72.5\%\\ (75.4\%)} &  \tabincell{c}{48.0\%\\ (51.5\%)} \\
People's Daily &   \tabincell{c}{50658\\ (4620)} & \tabincell{c}{2169879\\ (172590)} & \tabincell{c}{3\\ (3)} &  \tabincell{c}{42.83\\ (37.35)} & \tabincell{c}{1.47\\ (1.33)} &    \tabincell{c}{3.23\\ (3.25)} & \tabincell{c}{11.1\%\\ (11.6\%)} & \tabincell{c}{58.3\%\\ (54.4\%)} &  \tabincell{c}{35.8\%\\ (29.1\%)} \\
\textbf{CONLL2003} &   \tabincell{c}{13862\\ (3235)} & \tabincell{c}{203442\\ (51347)} & \tabincell{c}{4\\ (4)} &    \tabincell{c}{14.67\\ (15.87)} & \tabincell{c}{1.69\\ (1.83)} &    \tabincell{c}{1.44\\ (1.44)} & \tabincell{c}{16.7\%\\ (16.7\%)} & \tabincell{c}{79.9\%\\ (80.4\%)} &  \tabincell{c}{44.2\%\\ (48.8\%)} \\
\textbf{Ritter} &   \tabincell{c}{1955\\ (438)} & \tabincell{c}{37735\\ (8733)} & \tabincell{c}{10\\ (10)} &    \tabincell{c}{19.30\\ (19.93)} & \tabincell{c}{0.62\\ (0.60)} &    \tabincell{c}{1.65\\ (1.62)} & \tabincell{c}{5.3\%\\ (4.9\%)} & \tabincell{c}{38.1\%\\ (39.2\%)} &  \tabincell{c}{15.3\%\\ (15.5\%)} \\
\hline
\end{tabular}
\end{table*}

\subsection{Experimental Setting}
For each dataset, we random choose $1\%$ warmstart  samples as initial training set $\mathcal{L}_1$. We train initial model on this data, then we apply active learning strategy to choose additional $2\%$ samples based on model's uncertainty estimates and train a new model based on this data. In each iteration, we train from scratch to avoid negative effects accumulated from previous training. We train each model to convergence in each iteration. We fix the number of active learning iterations at $25$ because of each algorithm does not improve obviously after $25$ iteration.

In the NER model, we use a $300d$ word embedding pre-trained on the Chinese Wikipedia corpus\cite{P18-2023} for the Chinese datasets, and a $100d$ glove word embedding pre-trained on the English Wikipedia corpus\cite{pennington2014glove} for the English datasets. We uniformly set the learning rate as $0.001$ and the training batch size as $64$. The transition matrix $A$ in CRF is left to let the model learn by itself. It must be noted that the goal of this article is not to obtain SOTA of NER, but to compare the performance of different active learning strategies under same conditions. So, the NER model itself and its parameters may not be the best but fair.

We empirically compare the selection strategy proposed in Section \ref{sec:strategies}, as well as the uniformly random baseline (\textbf{RAND}) and long baseline (\textbf{LONG}). We evaluate each selection strategy by constructing learning curves that plot the overall $F_1$-score (for entities) and $accuracy$ (for sentences). In order to prevent the contingency of experiments, we have done $5$ independent experiments for each selection strategy on each dataset using different random initial training set $\mathcal{L}_1$. All results reported in this paper are averaged across these experiments.

\subsection{Results}

\begin{figure*}[!hbpt]
\centering
\subfloat[Boson\_NER]{\includegraphics[width=0.33\linewidth]{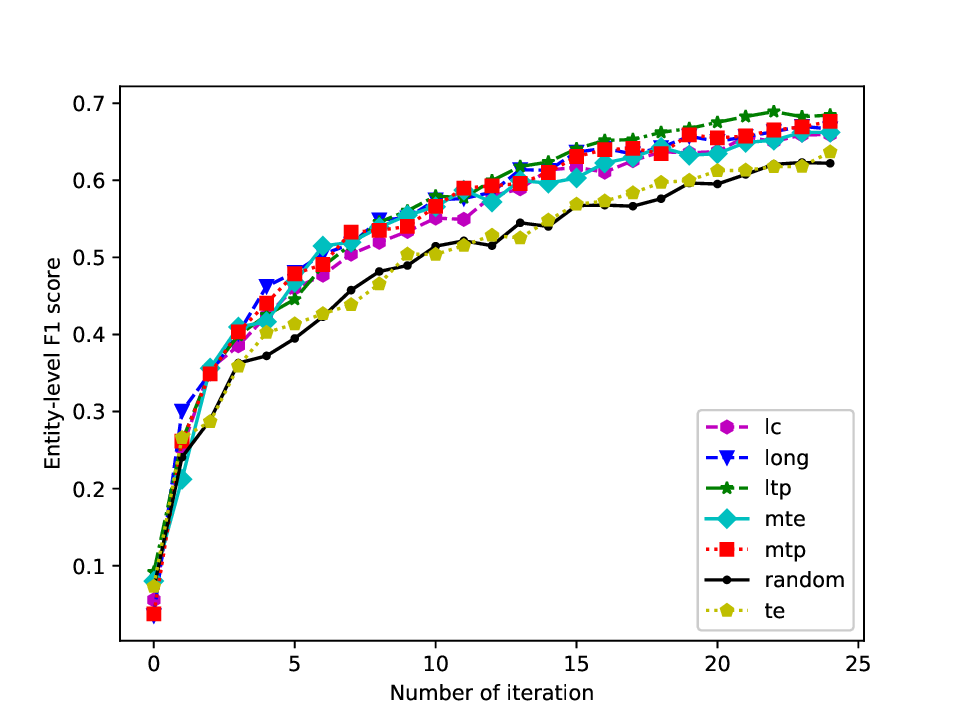}}\hfil
\subfloat[Weibo\_NER]{\includegraphics[width=0.33\linewidth]{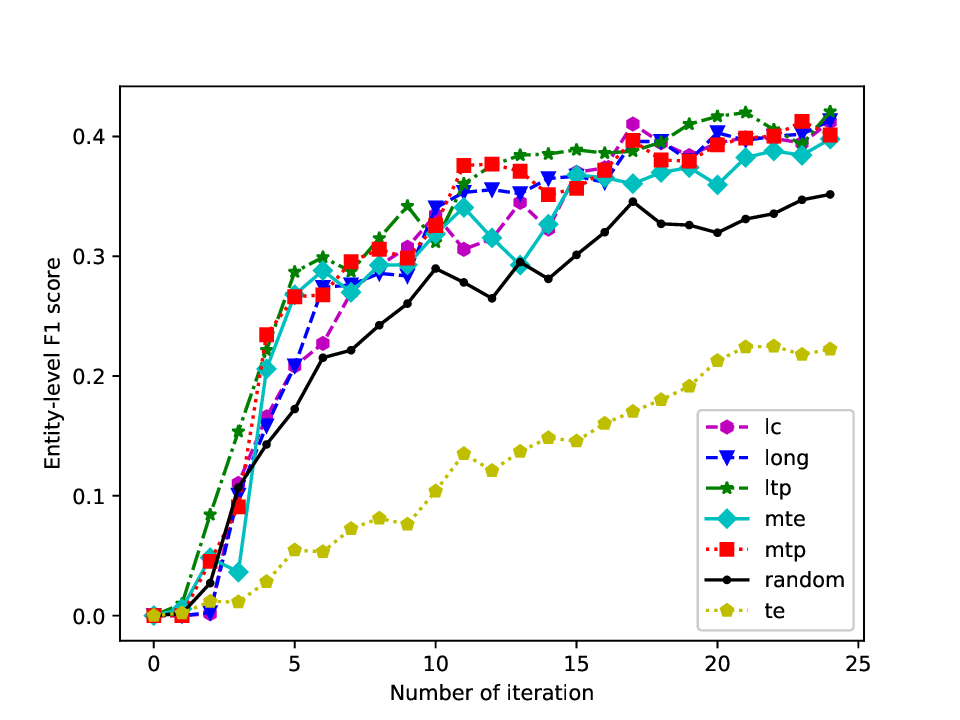}}\hfil
\subfloat[OntoNotes5.0]{\includegraphics[width=0.33\linewidth]{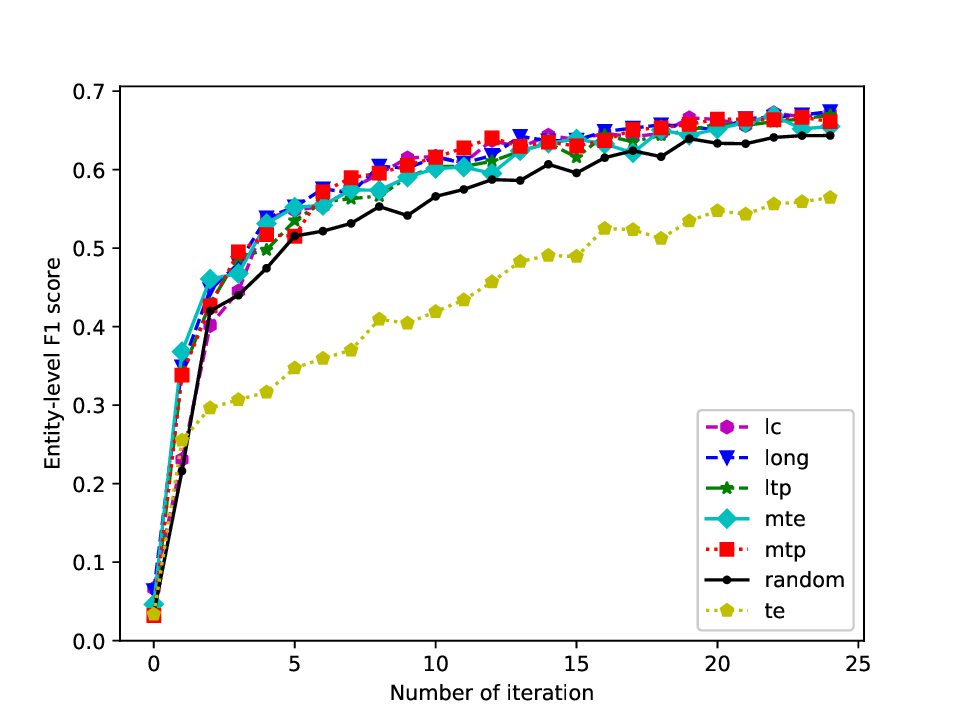}}

\subfloat[People's Daily]{\includegraphics[width=0.33\linewidth]{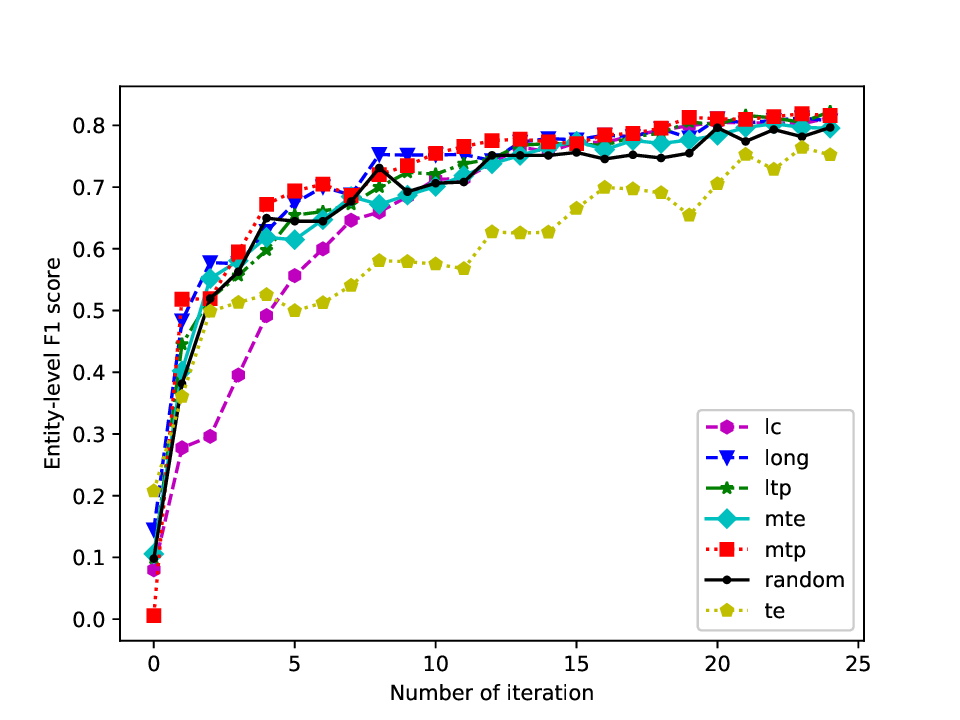}}\hfil
\subfloat[CONLL2003]{\includegraphics[width=0.33\linewidth]{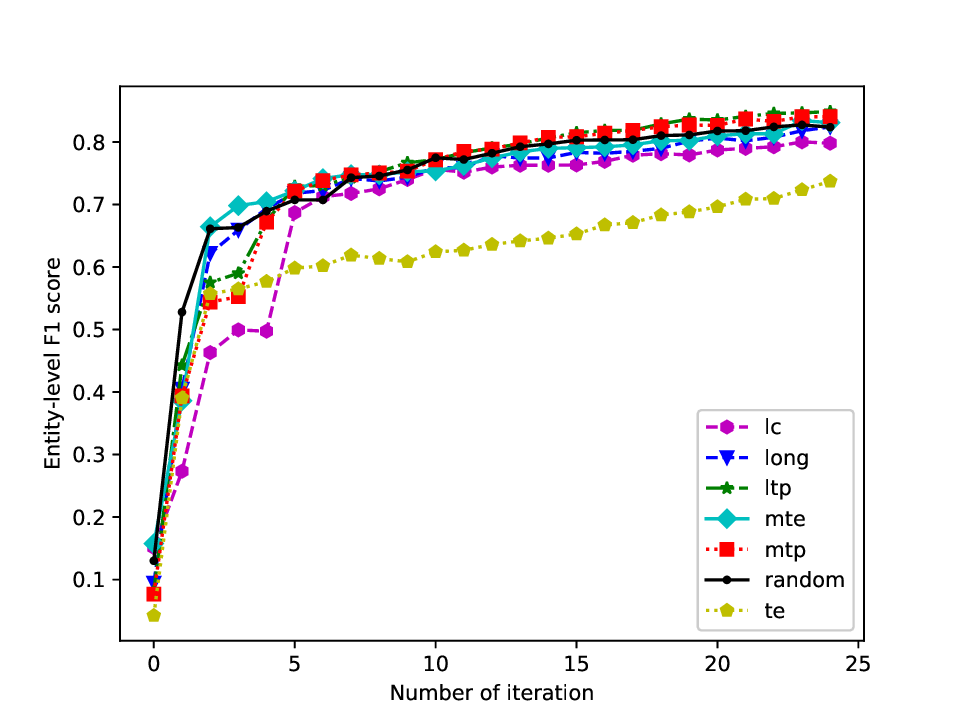}}\hfil
\subfloat[Ritter]{\includegraphics[width=0.33\linewidth]{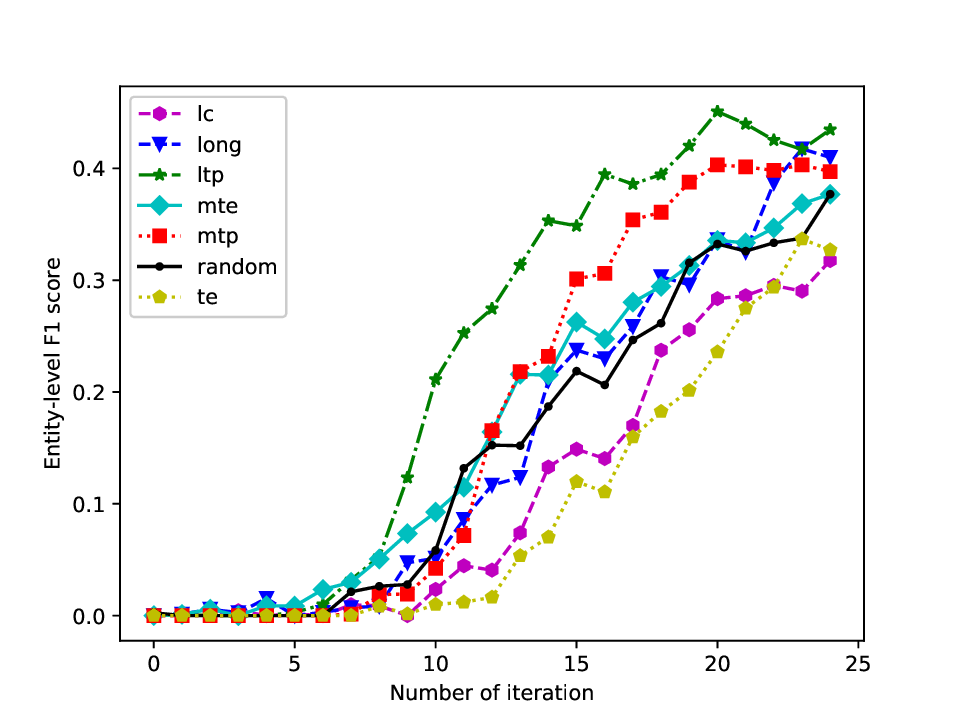}}

\caption{Entity-level $F_1$-score results on different datasets}\label{fig:f1}
\end{figure*}

\textbf{Entity-level $F_1$-scores} are shown in Figure \ref{fig:f1}, it is clear that all active learning strategies (except \textbf{TE}) perform better than the random baseline on $4$ Chinese datasets. Our strategy is not weaker than other strategies on all datasets, slightly better than other strategies on \textit{Boson\_NER}, \textit{Weibo\_NER}, and \textit{CONLL2003}, and significantly surpasses other strategies on \textit{Ritter}.

\begin{figure}[htbp]
	\centering
	\includegraphics[width=0.8\linewidth]{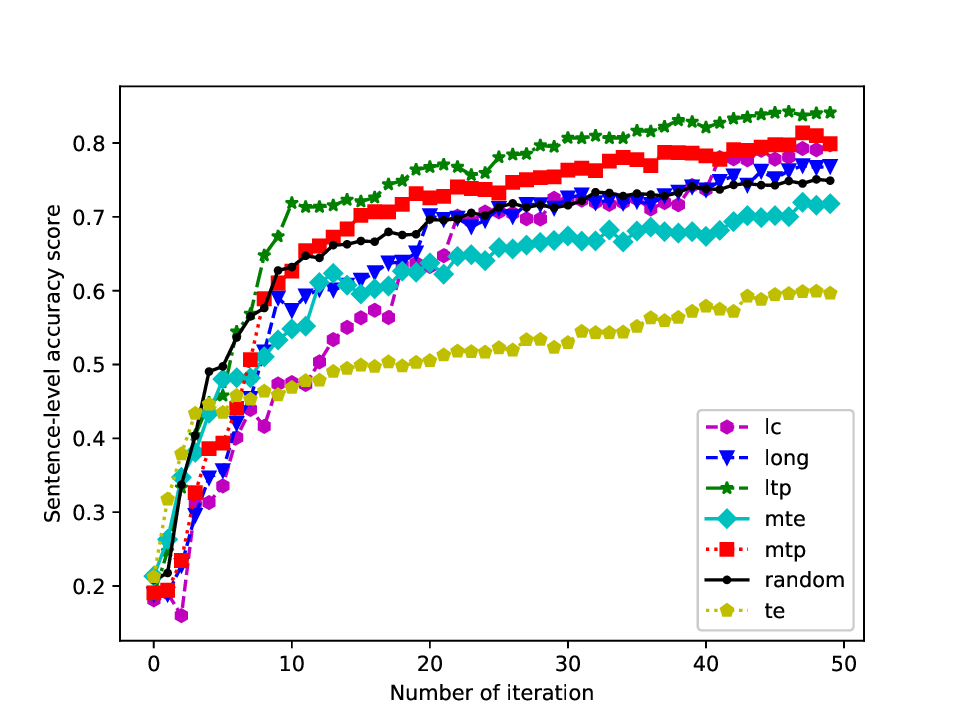}
	\caption{Sentence-level accuracy score results on CONLL2003 with $1\%$ samples selected each iteration.}\label{fig:conll-1}
\end{figure}

\begin{figure*}[!htbp]
\centering
\subfloat[Boson\_NER]{\includegraphics[width=0.33\linewidth]{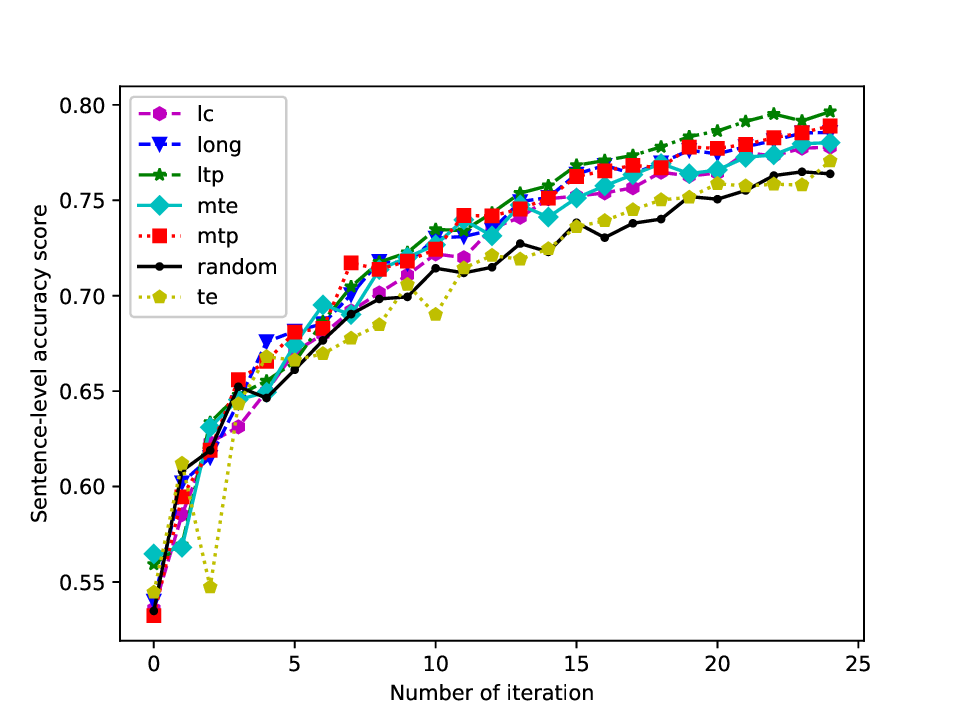}}\hfil
\subfloat[Weibo\_NER]{\includegraphics[width=0.33\linewidth]{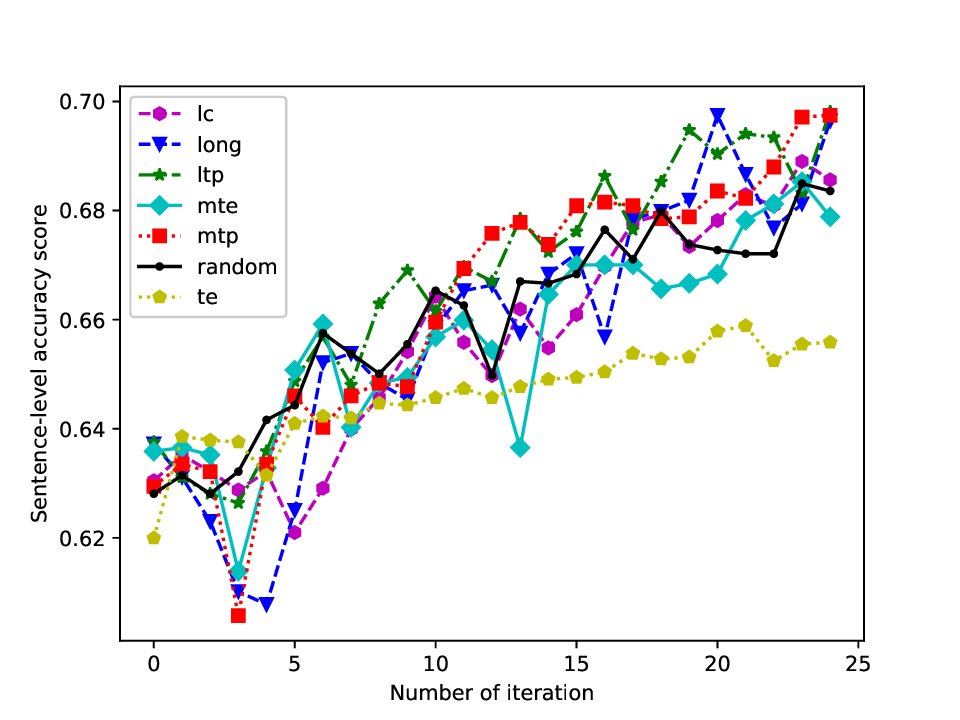}}\hfil
\subfloat[OntoNotes5.0]{\includegraphics[width=0.33\linewidth]{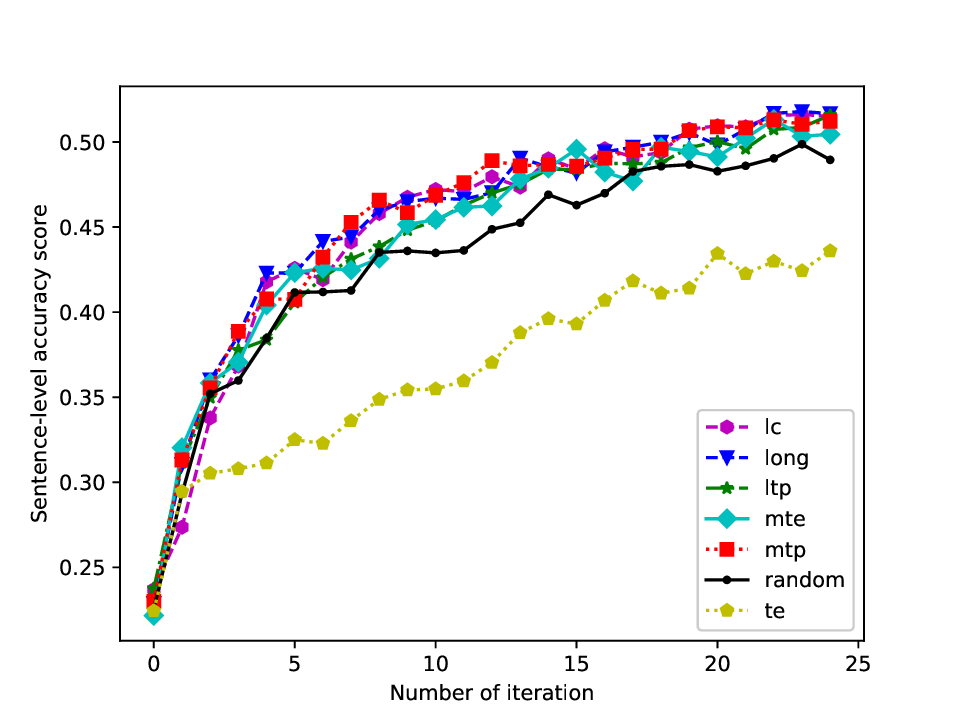}}

\subfloat[People's Daily]{\includegraphics[width=0.33\linewidth]{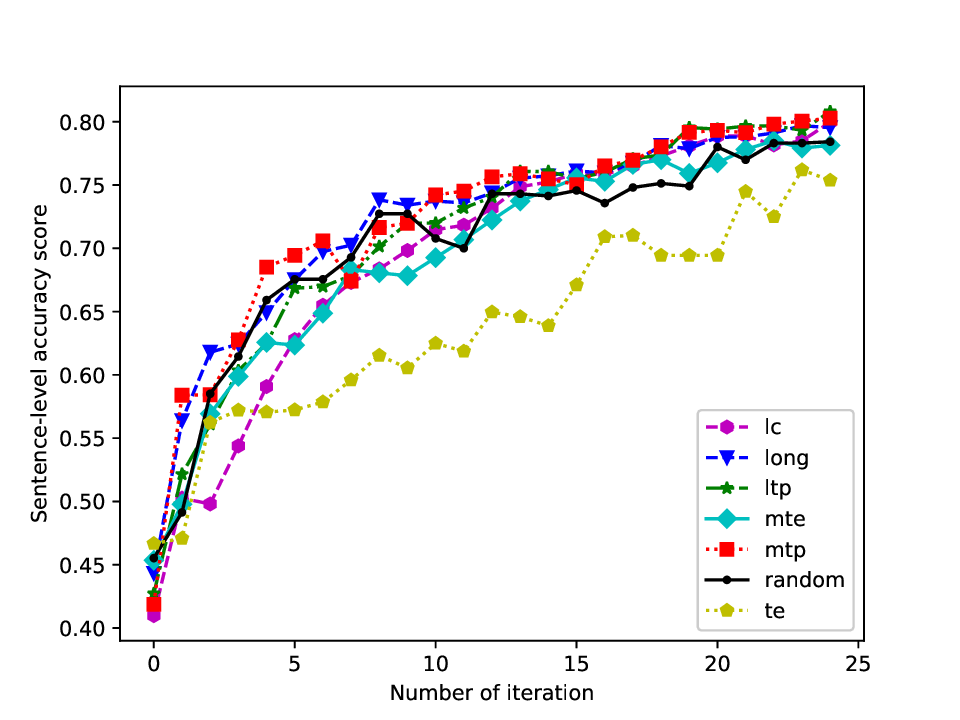}}\hfil
\subfloat[CONLL2003]{\includegraphics[width=0.33\linewidth]{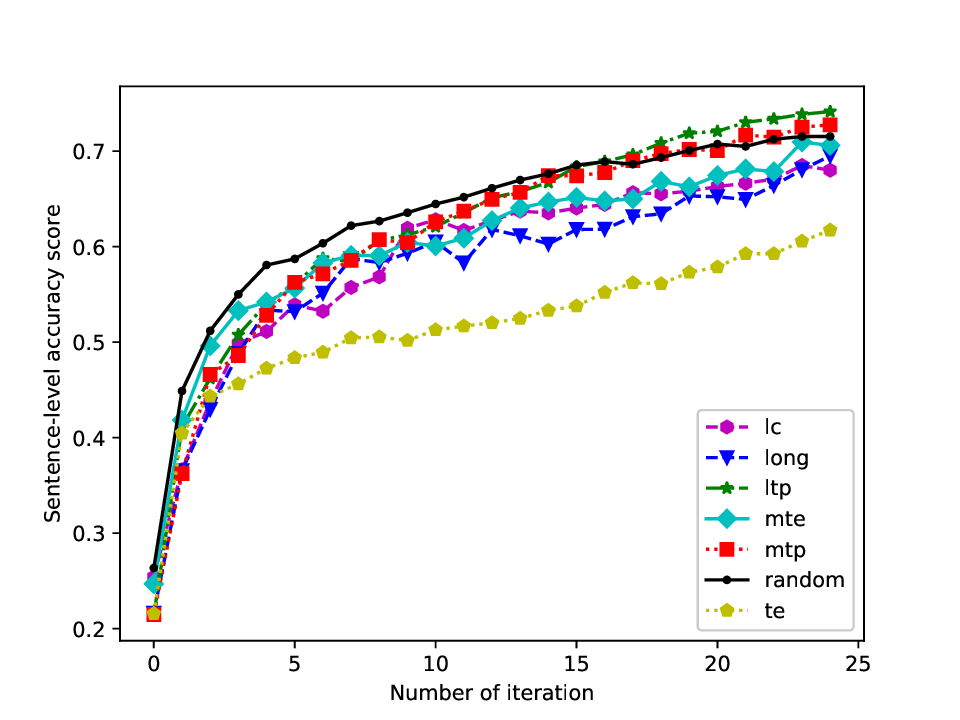}}\hfil
\subfloat[Ritter]{\includegraphics[width=0.33\linewidth]{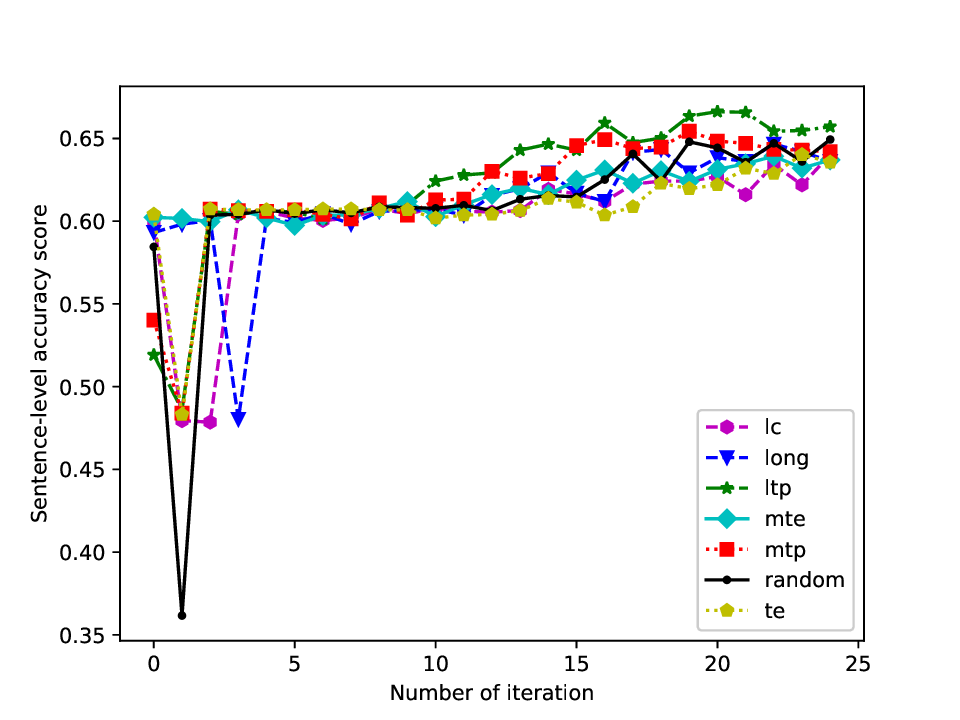}}

\caption{Sentence-level accuracy score results on different datasets}\label{fig:acc}
\end{figure*}

Figure \ref{fig:acc} shows the results of \textbf{sentence-level accuracy} on six datasets. The results exceeded our expectations and are very interesting. Firstly, the results confirm that entity-level $F_1$-score is sometimes misleading (two social media datasets, \textit{Weibo\_NER} and \textit{Ritter}) as what we mention in Section \ref{sec:intro}. Secondly, our strategy \textbf{LTP} is better than the rest of methods, while it not obvious on the large data set of canonical text, which is similar to text for pre-trained word embedding.


\begin{figure*}[htbp]
\centering
\subfloat[Boson\_NER]{\includegraphics[width=0.33\linewidth]{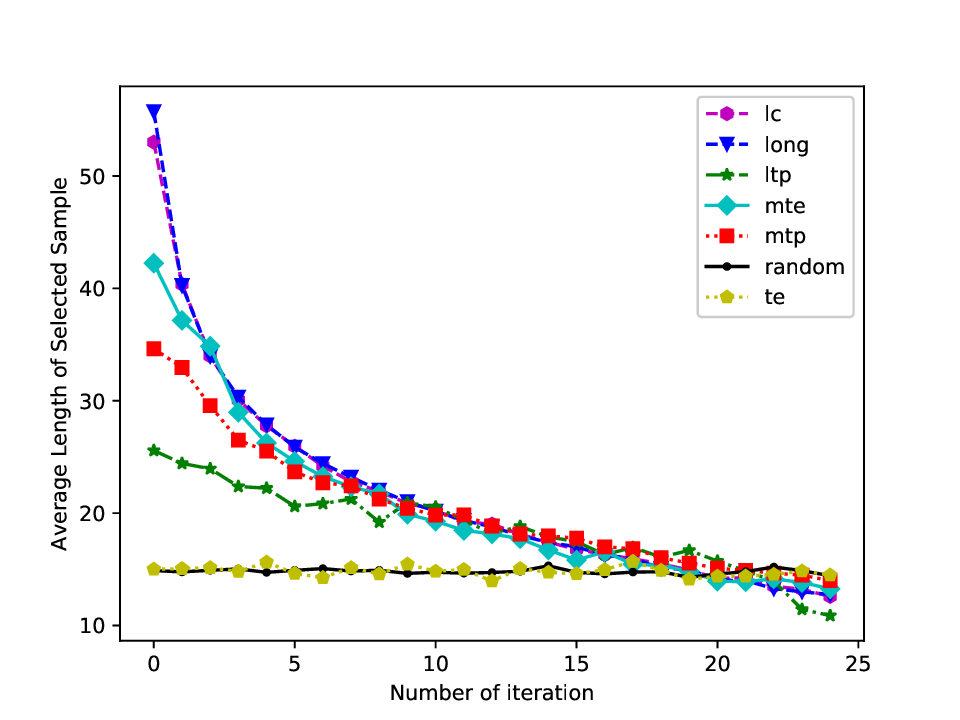}}\hfil
\subfloat[Weibo\_NER]{\includegraphics[width=0.33\linewidth]{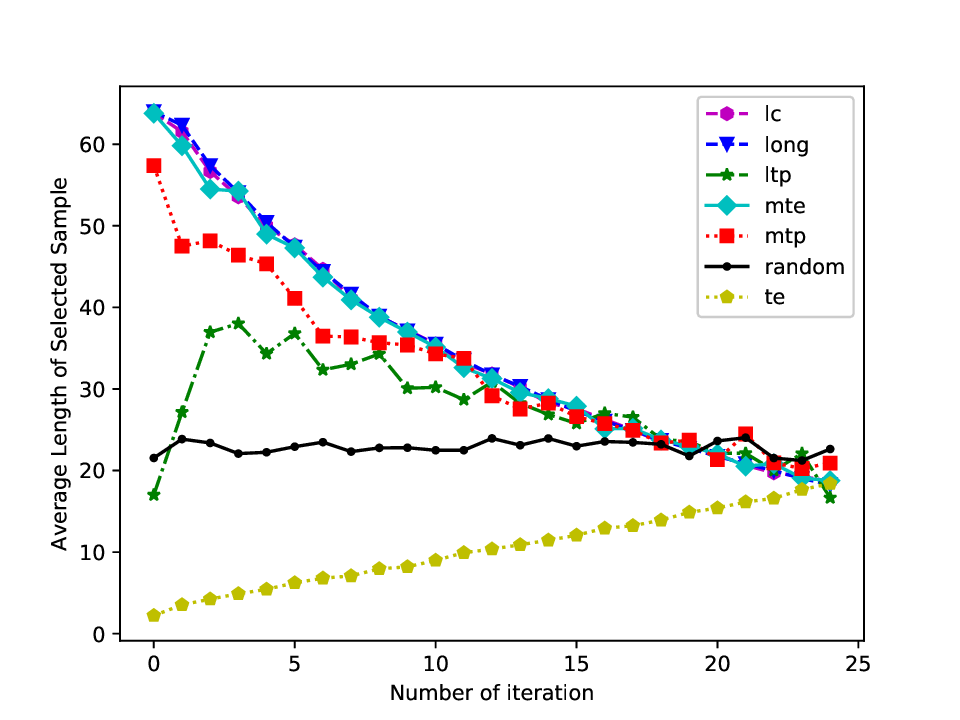}}\hfil 
\subfloat[OntoNotes5.0]{\includegraphics[width=0.33\linewidth]{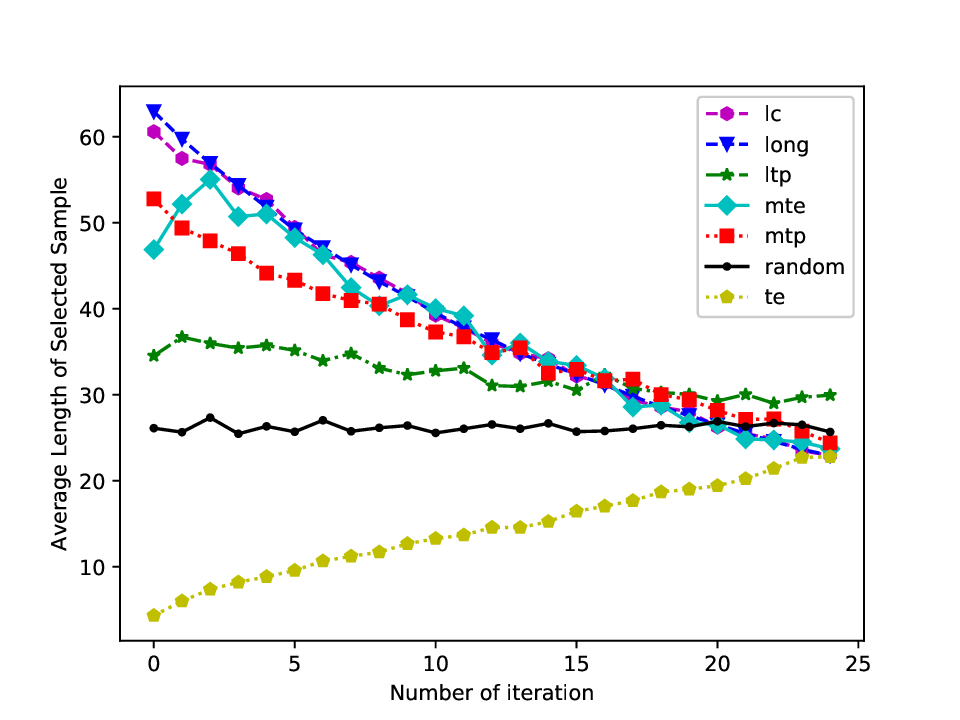}}

\subfloat[People's Daily]{\includegraphics[width=0.33\linewidth]{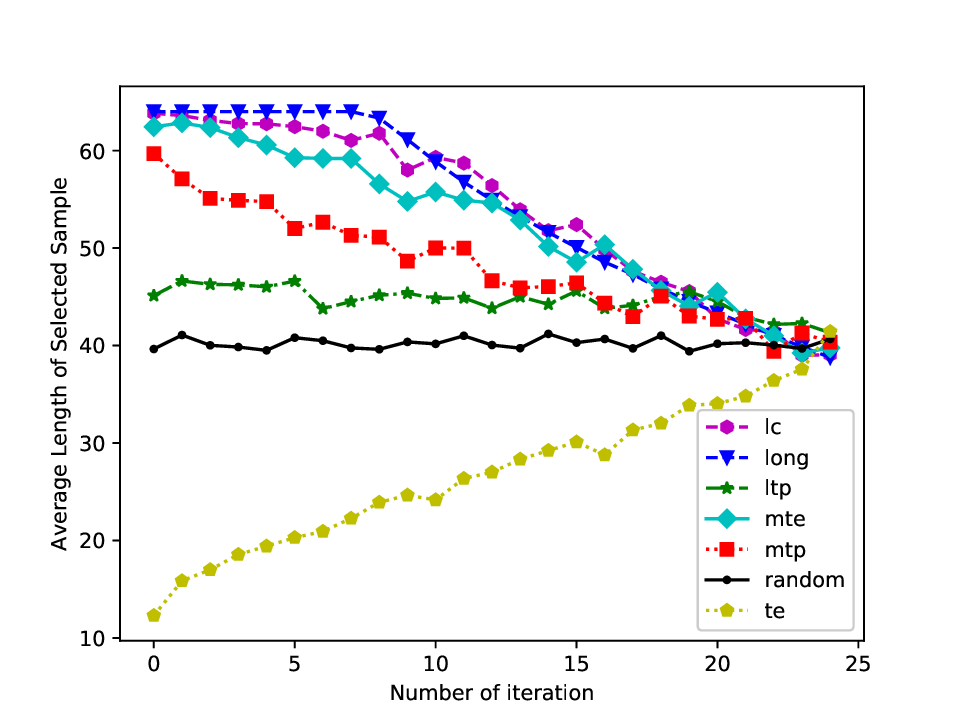}}\hfil
\subfloat[CONLL2003]{\includegraphics[width=0.33\linewidth]{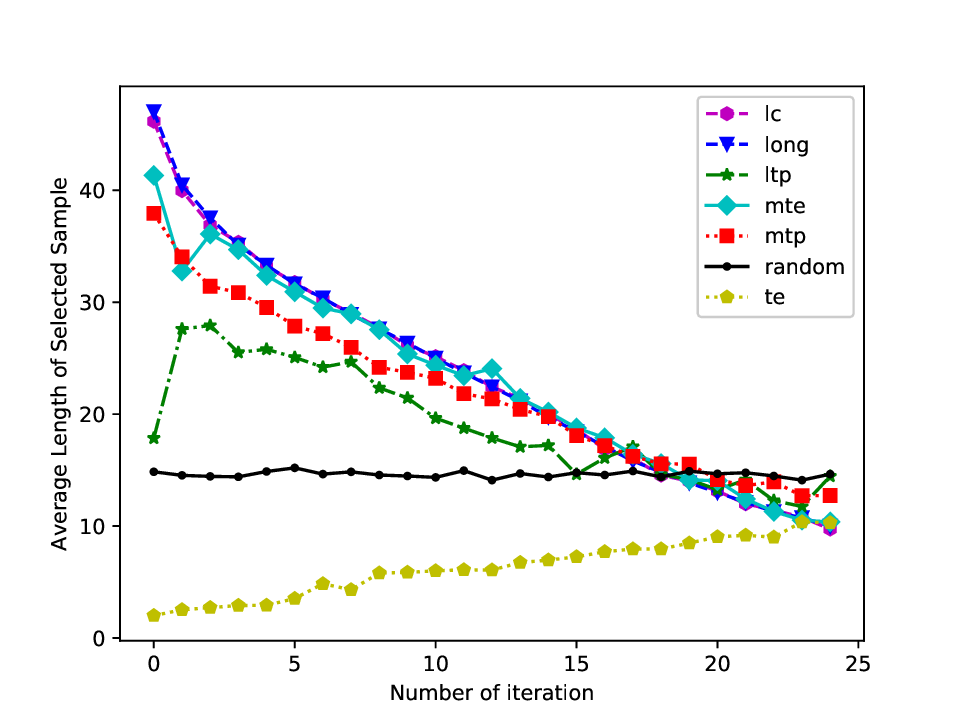}}\hfil
\subfloat[Ritter]{\includegraphics[width=0.33\linewidth]{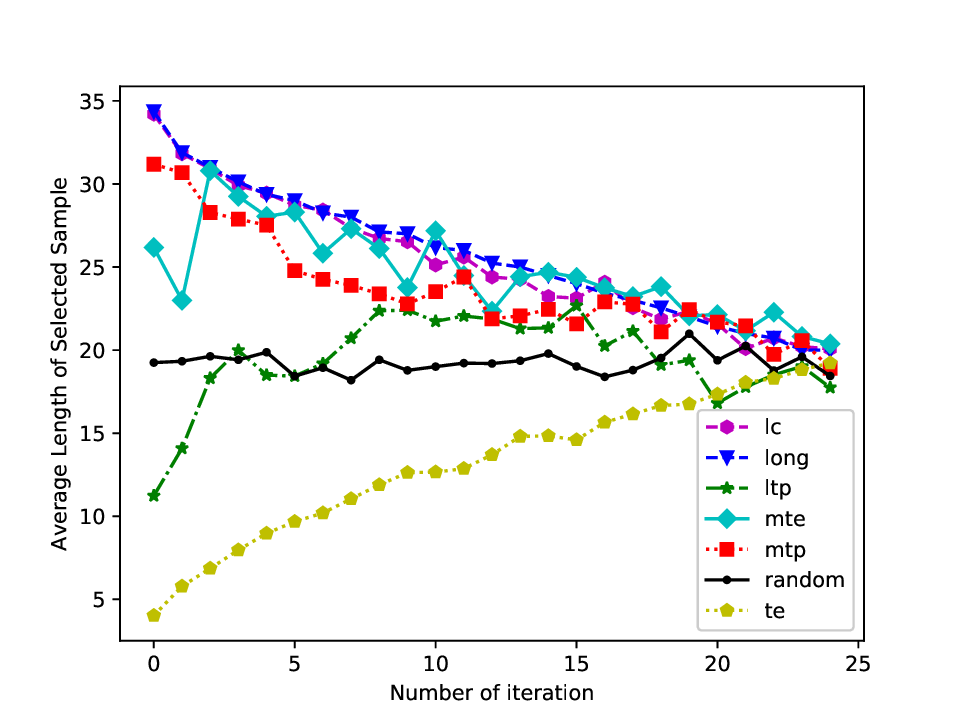}}

\caption{Average length of the samples selected by active learning strategies.}\label{fig:length}
\end{figure*}

\begin{figure*}[htbp]
\centering
\subfloat[Boson\_NER]{\includegraphics[width=0.33\linewidth]{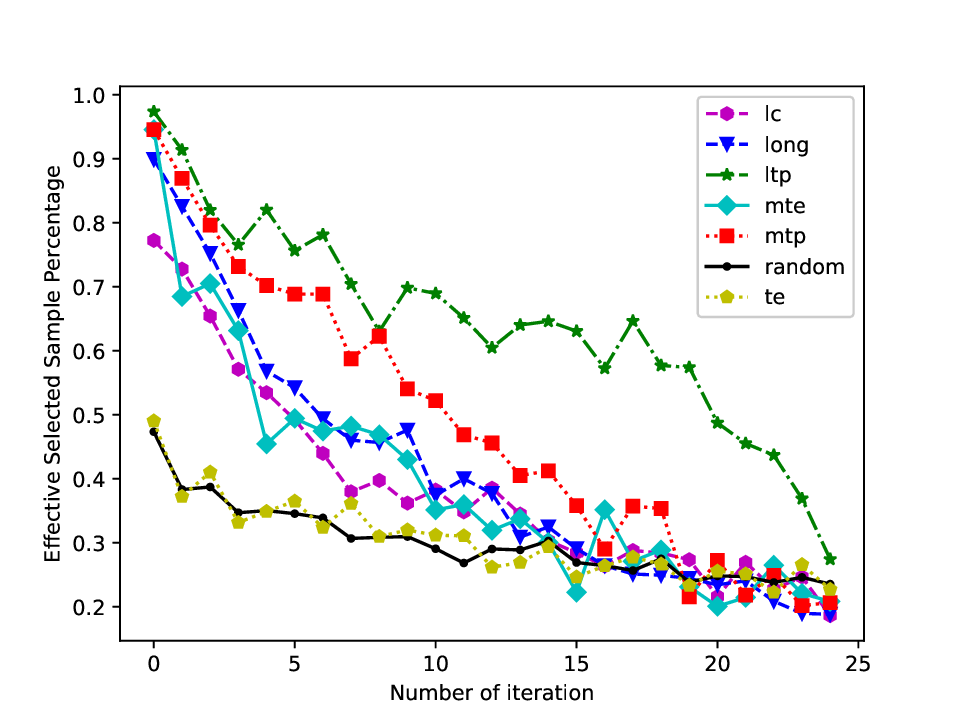}}\hfil
\subfloat[Weibo\_NER]{\includegraphics[width=0.33\linewidth]{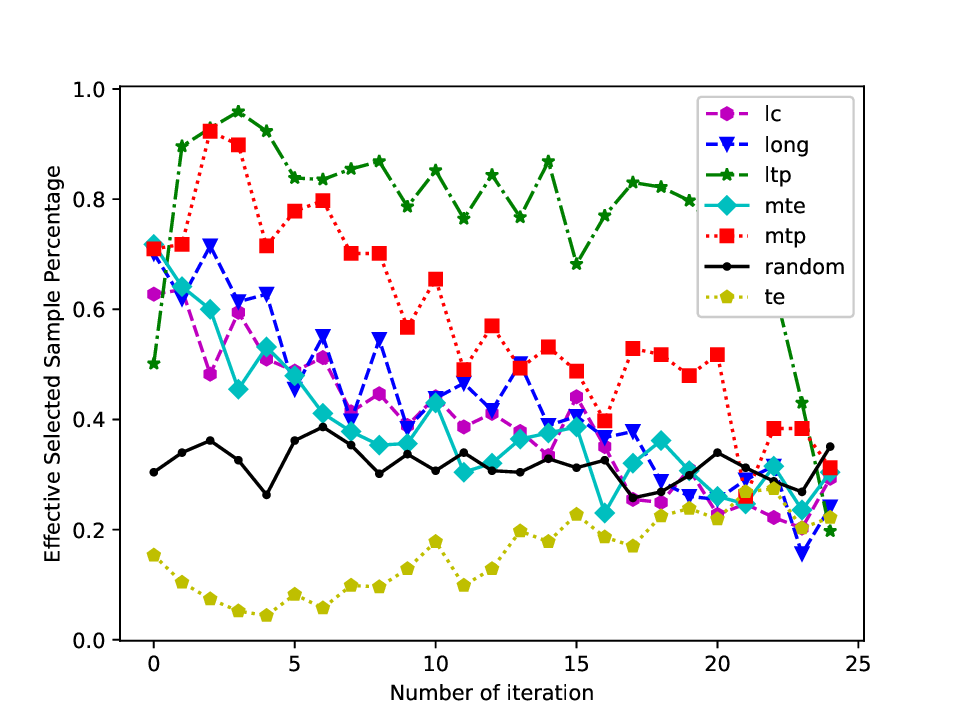}}\hfil
\subfloat[OntoNotes5.0]{\includegraphics[width=0.33\linewidth]{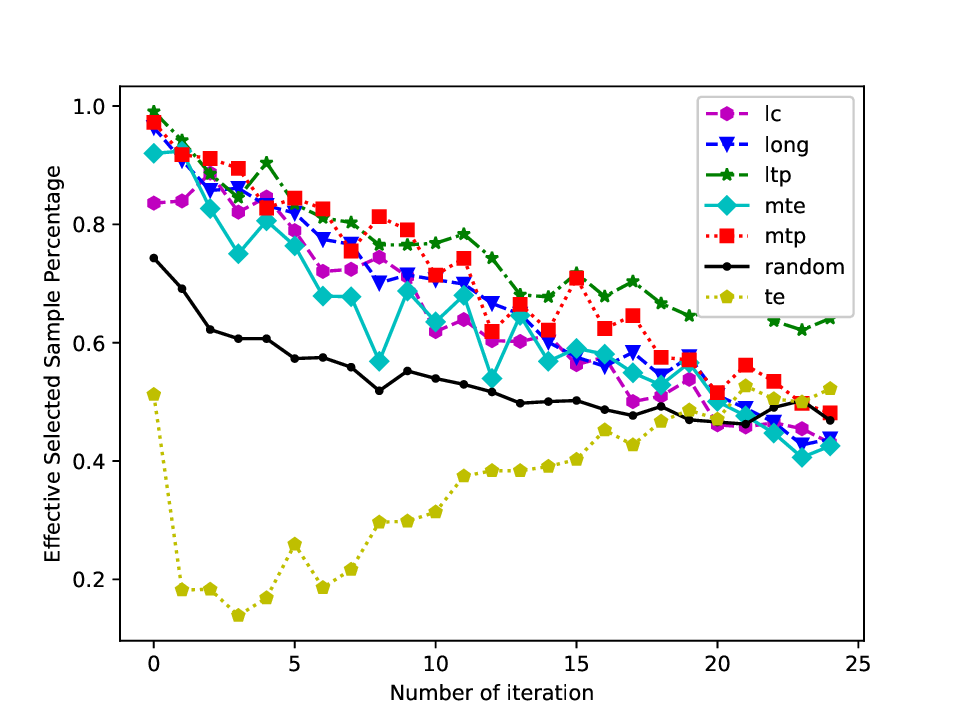}}

\subfloat[People's Daily]{\includegraphics[width=0.33\linewidth]{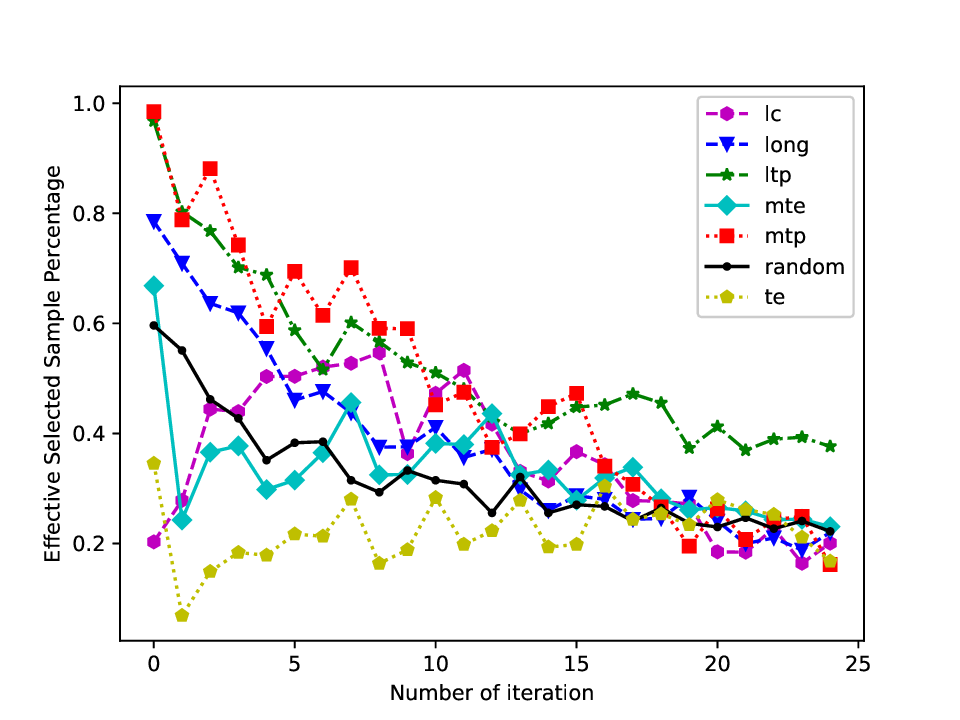}} \hfil
\subfloat[CONLL2003]{\includegraphics[width=0.33\linewidth]{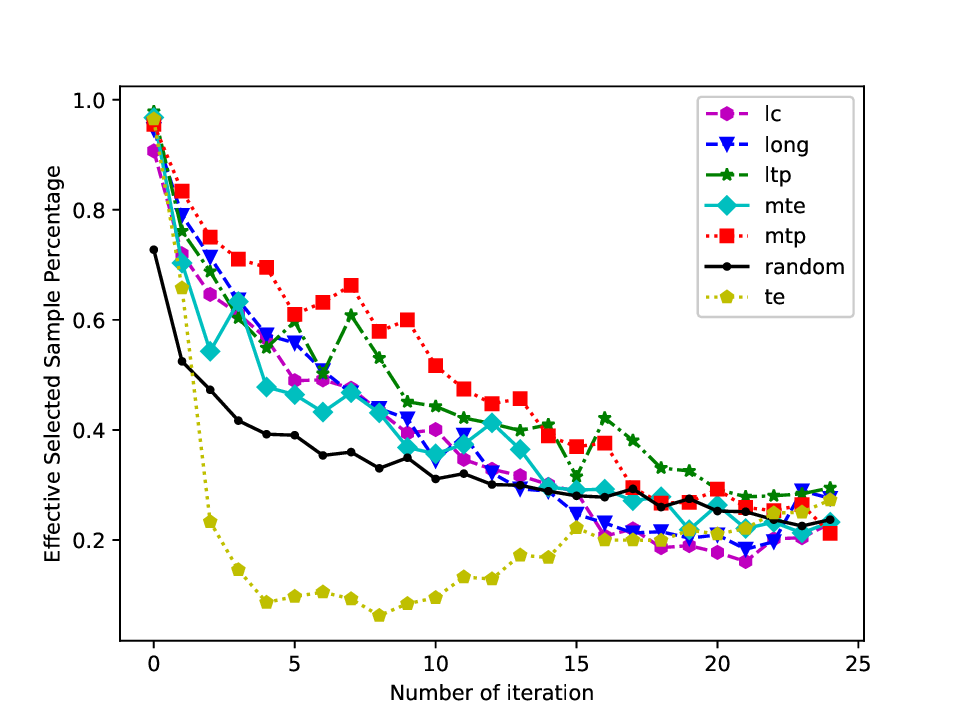}}\hfil
\subfloat[Ritter]{\includegraphics[width=0.33\linewidth]{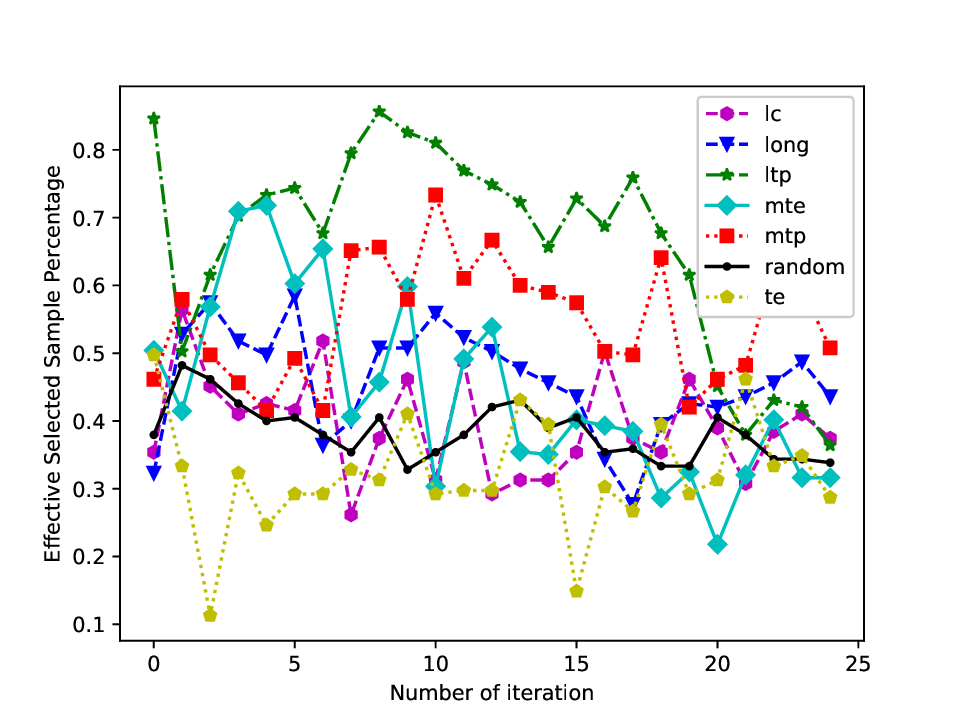}}

\caption{The results of effective selected sample percentage on different datasets}\label{fig:effectiveness}
\end{figure*}
We known that the most obvious effect of active learning is to select one sample at a time, although this is not realistic due to the cost of retraining. The more samples selected each time, the worse the active learning effect. Therefore, in the case of a large data pool, selecting $2\%$ of the samples in each round cannot clearly reflect the differences between different strategies. In order to clearly reflect the differences between strategies, we constructed an additional experiment on \textit{CONLL2003} with \textbf{$1\%$} samples selected each iteration. Results are given in Figure \ref{fig:conll-1}.

Figure \ref{fig:length} shows \textbf{average length} of the samples selected by different active learning strategies. Unlike other active learning strategies, \textbf{LTP} does not have a obvious bias towards sample length. From another aspect, \textbf{LTP} use less annotation cost to achieve better performance than other strategies.

\subsection{Discussion}
In this section, we will briefly discuss \textit{possible reasons} for the gap between different selection strategies. 

The core of active learning is to select "informative" samples, but there is no unified standard to measure "informative". One thing is certain, the samples that are not correctly labeled by the model are informative samples for the model. Therefore, we use the proportion of samples in each iteration of selection the model is not correctly labeled as the effectiveness of each iteration of selection. Figure \ref{fig:effectiveness} shows the results. We	 can find that \textbf{LTP} can more effectively select samples that are incorrectly predicted by the model.

\section{Conclusion}
We proposed a new active learning strategy for CRF-based named entity recognition. The experiment shows that compared with the traditional active selection strategies, our strategy has better performance, but lower annotation cost.

\section{Acknowledgments}
Research in this paper is partially supported by the National Key Research and Development Program of China (No 2018YFB1402500), the National Science Foundation of China (61832004, 61772155, 61802089, 61832014).


\bibliographystyle{elsarticle-num} 
\bibliography{main.bib}


\end{document}